\documentclass[12pt]{article}
\usepackage[utf8]{inputenc}

\usepackage{adjustbox}
\usepackage[scaled=1]{helvet}
\usepackage[helvet]{sfmath}
\everymath={\sf}
\usepackage[labelfont=bf]{caption}
\usepackage{float}
\usepackage[onehalfspacing]{setspace}
\usepackage[margin=2.5cm]{geometry}
\usepackage{amsfonts}
 \usepackage{adjustbox}
\usepackage{mathtools, nccmath}
\usepackage{pdfpages}

\makeatletter
\renewcommand{\paragraph}{%
  \@startsection{paragraph}{4}%
  {\z@}{1.5ex \@plus 1ex \@minus .2ex}{-1em}%
  {\normalfont\normalsize\bfseries}%
}
\makeatother

\usepackage[hidelinks]{hyperref}
\usepackage[numbers]{natbib}
\usepackage{comment}
\usepackage{changepage}
\usepackage{nicematrix}
\usepackage[T1]{fontenc}
\usepackage{tikz}
\DeclareFontFamily{OMS}{lmtt}{} % xx todo added missing font to squelch warnings which is not rendered i believe - Jonathan
\DeclareFontShape{OMS}{lmtt}{m}{n}{<->ssub * cmsy/m/n}{}

\setcitestyle{square}
\usepackage[
singlelinecheck=false % <-- important
]{caption}
\usepackage{titling}
\newcommand{\subtitle}[1]{%
  \posttitle{%
    \par\end{center}
    \begin{center}\large#1\end{center}
    \vskip0.5em}%
}
\usepackage{xurl}
\usepackage{amsmath}
\usepackage{xcolor}
\usepackage{longtable}
\usepackage{wrapfig}
\usepackage{subcaption}
\usepackage{lmodern}
\usepackage{textcomp}
\usepackage{gensymb}
\usepackage{graphicx}
\usepackage{amsmath}
\usepackage{siunitx}
\usepackage{verbatim}

\usepackage{xcolor}         % colors
\usepackage{graphicx}
\usepackage{booktabs}
\usepackage{multirow}
\title{Concept-based explanation of gene expression prediction from H\&E images}

\author{%
  \begin{minipage}{\textwidth}
  \centering
  Amos Muench$^{1}$ \quad Jonathan Thielmann$^{1,2}$ \quad Reduan Achtibat$^{2}$ \\
  Maximilian Dreyer$^{2}$ \quad Philip Bischoff$^{2}$ \quad Caroline Forsythe$^{1}$ \\
  Hamidreza Parand$^{1}$ \quad Thomas Walter$^{3,4}$ \quad David Horst$^{1,5}$ \\
  Sebastian Lapuschkin$^{2,6}$ \quad Wojciech Samek$^{2,7,8}$ \quad Teresa Gabriela Krieger$^{1,5}$ \\[1ex]
  $^{1}$ Institute of Pathology -- Charité Universitätsmedizin, Berlin, Germany \\
  $^{2}$ Fraunhofer Heinrich-Hertz-Institute, Berlin, Germany \\
  $^{3}$ Center for Computational Biology (CBIO), Mines Paris, PSL University, Paris, France \\
  $^{4}$ Institut Curie, PSL University, Paris, France \\
  $^{5}$ German Cancer Consortium (DKTK), Partner Site Berlin, and German Cancer Research Center (DKFZ), Heidelberg, Germany \\
  $^{6}$ Technological University Dublin, Dublin, Ireland \\
  $^{7}$ Technische Universität Berlin, Berlin, Germany \\
  $^{8}$ BIFOLD -- Berlin Institute for the Foundations of Learning and Data, Berlin, Germany \\[1ex]
  Correspondence to: Amos Muench, \texttt{amos.muench@charite.de}
  \end{minipage}
}
\begin{document}

\maketitle

\clearpage
\begin{abstract}
\noindent\textbf{Background}\\
Recent advances in pathology foundation models have enabled accurate prediction of spatial transcriptomics (ST) from routine H\&E images. However, existing explainability methods for vision transformer (ViT)-based models are largely limited to local heatmaps and do not reveal how morphological concepts contribute to ST predictions. Here, we introduce an explainable framework that combines relevance propagation and concept discovery to link transcriptional programs to tissue morphology.\\
\textbf{Methods}\\
We developed a ViT-based framework for virtual ST from H\&E images that combines ViT-aware layer-wise relevance propagation with relaxed archetypal TopK sparse autoencoder-based concept discovery. This approach provides both local explanations and global insights into the morphological patterns associated with transcriptional programs. We applied the framework to colorectal cancer ST data from the HEST-1k cohort and evaluated its generalizability in TCGA COAD.\\
\textbf{Results}\\
Our architecture accurately predicts clinically relevant ST signatures and accompanying molecular phenotypes. Measured and predicted gene expression profiles reveal substantial spatial heterogeneity of the colorectal cancer subtypes iCMS2 and iCMS3 across a large number of samples. Spatially resolved and aggregated iCMS classification achieve weighted F1 scores of 0.872 and 0.819 (0.770 in TCGA COAD), respectively, and both stratify patient outcome. Beyond prediction, our framework establishes a relevance-based concept atlas linking molecular phenotypes to histopathological representations. Comparison of activation- with relevance-derived concepts demonstrates that relevances provide a more direct link between tissue morphology and downstream predictions.\\
\textbf{Conclusions}\\
We establish a general strategy for concept-based explanation of spatial prediction, and our framework is readily applicable to a broad range of ViT-based pathology models.
\end{abstract}
\clearpage

\section{Introduction}

Routine histopathology slides stained with hematoxylin and eosin (H\&E) remain central to cancer diagnosis and are generated for nearly all patients undergoing pathological evaluation. Deep learning in computational pathology can predict diverse patient-level characteristics from whole-slide images (WSIs), including mutational status, molecular subtype, treatment response, and survival \cite{Campanella2019Clinical,Coudray2018ClassificationLearning,Kather2019DeepCancer}. However, many clinically relevant phenotypes arise from spatially heterogeneous biological processes within distinct tissue regions. This intra-tumoral heterogeneity, a key determinant of disease progression, treatment resistance, and outcome \cite{DagogoJack2018TumourTherapies,Li2022UntanglingHeterogeneity,Roerden2025CancerHeterogeneity}, is only partially captured by slide-level prediction.

Spatial transcriptomics (ST) enables direct characterization of molecular heterogeneity in intact tissues \cite{Ke2013InCells, Stahl2016VisualizationTranscriptomics}, preserving both histological context and spatially resolved GE. With increasing availability of paired H\&E and ST datasets, recent work has focused on predicting spatial GE directly from routine histology. Foundation-model (FM) frameworks such as BLEEP \cite{Xie2023SpatiallyLearning} and Phoenix \cite{Tran2026PancancerPhoenix} have substantially improved predictive performance and cross-tissue generalizability, facilitating the construction of virtual ST atlases \cite{Nonchev2026DeepSpot-M:Histology}. Among these, vision transformers (ViTs) are particularly promising because they capture long-range contextual relationships central to tissue organization \cite{Dosovitskiy2021ViT}. However, almost all digital pathology ViT FMs use small image tiles as input, thereby ignoring long range interactions across WSIs, but still allowing for attention across all subregions of small image tiles.
Despite rapid progress, digital pathology FMs still lag behind other imaging benchmarks, likely due to architectures limited to small image patches that fail to capture larger mesostructures, and to object patterns that are poorly aligned with self-supervised learning paradigms \cite{WhyFailing}.

Existing interpretability methods in computational pathology have largely been developed for slide-level prediction tasks, where explanations are typically linked to global patient labels \cite{Lu2021CLAM,Hense2024xMIL}. Recent work by Jamshidi Idaji et al. \cite{JamshidiIdaji2026BeyondHistopathology} has analyzed the faithfulness of heatmaps generated by different explanatory frameworks and found that perturbation-based, layer-wise relevance propagation (LRP), and integrated gradients consistently outperformed attention-based and gradient-based saliency heatmaps on multiple tasks predicting bulk information from WSIs. However, molecular states frequently vary within a single tissue section, making local and spatially resolved explanations essential for understanding model behavior and generating biologically meaningful insights. This challenge is particularly pronounced for GE prediction, where models learn complex relationships between morphology and transcriptional programs that may lack an obvious histological correlate.

LRP provides a principled framework for attributing predictions to (image) inputs and has become one of the most widely used methods for spatial explanation of deep neural networks \cite{Bach2015LRP}. However, local relevance maps alone are often noisy and difficult to interpret, particularly in transformer architectures whose predictions emerge from distributed representations. Concept-based interpretability methods offer a complementary perspective by linking model decisions to interpretable latent features. In particular, concept relevance propagation (CRP) enables relevance to be traced through learned latent concepts, thereby connecting local explanations with global patterns of model reasoning \cite{Achtibat2023CRP, Dreyer2024PCX}. 
Nevertheless, concept-based explanation frameworks have not yet been explored for GE prediction. Existing heatmap-based approaches localize predictions but do not explain them in terms of higher-level morphological concepts, whereas concept-discovery methods identify latent concepts but are typically disconnected from spatially resolved molecular prediction. Our framework aims to bridge these two perspectives.

Recent advances in sparse autoencoders have demonstrated that semantically meaningful concepts can be extracted from the latent representations of large vision models by decomposing activations into an overcomplete dictionary of features \cite{Elhage2022ToySuperposition,Park2024TheModels}. This approach is motivated by the linear representation hypothesis, which posits that neural networks encode information as a superposition of latent features that can be approximately recovered through sparse dictionary learning. However, recent theoretical and empirical work by Fel et al. \cite{Fel2025IntoGeometry} challenges the view that concepts correspond to sparse, orthogonal directions in activation space. Instead, they show that concept representations can exhibit richer geometric structure, including dense and overlapping features as well as antipodal concepts encoding redundant information. Moreover, the geometry and interpretability of concepts appear to depend strongly on the underlying task. To date, these observations have been studied primarily in DINOv2 representations trained and evaluated on ImageNet, mostly for classification tasks (and, in the case of regression, solely for depth-of-field
estimation), with a focus on analyzing activation spaces rather than attributing downstream predictions [16]. Existing concept-discovery approaches thus provide little insight into how latent concepts contribute to downstream predictions in transformer-based models, particularly in spatial regression tasks. Whether the geometric properties described by [16]  extend to spatial gene-expression prediction models and whether activation-derived concepts can support faithful explanations of molecular predictions remains unknown.
To address these challenges, we extend CRP to ViTs, enabling concept-level attribution in an architecture class for which existing CRP formulations are not directly applicable. Building on this adaptation, we develop an explainable framework for virtual spatial transcriptomics that combines accurate GE prediction with human-interpretable analysis of the morphological concepts associated with model predictions. Specifically, we integrate attention-aware layer-wise relevance propagation~\cite{Achtibat2024AttnLRP:Transformers} with a sparse autoencoder-based concept discovery framework that decomposes learned representations into interpretable concepts. Unlike prototype-based approaches that enforce predefined explanatory structures during training and may compromise predictive performance, our method derives concepts post hoc from unconstrained latent representations. This design preserves model flexibility while providing both local explanations of individual predictions and global insight into the morphological patterns associated with predicted transcriptional programs. Because the framework operates post hoc, it can be readily applied to diverse architectures and prediction tasks, including emerging pathology foundation models. It thereby establishes a general strategy for post hoc concept-based explanation of spatial prediction models. Our contributions are thus threefold: (i) we extend CRP to ViT architectures, (ii) we combine attention-aware relevance propagation with sparse autoencoder-based concept discovery for virtual spatial transcriptomics, and (iii) we demonstrate how this framework reveals morphological patterns associated with spatial transcriptional programs.

As a clinically relevant use case, we focus on colorectal cancer (CRC), a disease characterized by substantial molecular and spatial heterogeneity. Transcriptomic classification of CRC has evolved from the Consensus Molecular Subtypes (CMS1–4) derived from bulk RNA sequencing to more recent integrative frameworks incorporating genomic information \cite{Guinney2015TheCancer, Nunes2024PrognosticCancers}. However, single-cell transcriptomic analyses have shown that many bulk-derived classifications primarily reflect differences in cellular composition rather than intrinsic tumour cell states. Joanito and colleagues identified two malignant epithelial programs, termed intrinsic CMS2 (iCMS2) and intrinsic CMS3 (iCMS3), and proposed gene signatures capturing these tumour-specific transcriptional states \cite{Joanito2022scRNA}. These signatures stratify patient outcomes when applied to bulk transcriptomic cohorts, yet their spatial organization within tumours remains poorly understood despite emerging evidence that co-occurrence of distinct intrinsic programs may carry prognostic significance \cite{Langerud2024cCMS}. 

Here, we apply our framework to predict stromal, immune and intrinsic CRC transcriptional signatures directly from H\&E images, enabling spatial classification of iCMS states and investigation of their intra-tumoral organization. Simultaneously, we introduce a concept-based explanation framework that combines attention-aware relevance propagation with sparse autoencoder-based concept discovery to provide local and global insights into the morphological features associated with GE predictions. Beyond achieving state-of-the-art performance for clinically relevant transcriptomic signatures, our work establishes a general strategy for interpretable virtual spatial transcriptomics and, to our knowledge, represents one of the first applications of relaxed archetypal sparse autoencoder-based concept discovery in computational pathology.

\section{Methods}
\subsection{Datasets and preprocessing}
We retrieved the COAD subset of the HEST-1k dataset \cite{Jaume2024HEST-1k:Analysis} and kept n = 49 samples with a sufficient number of genes (omitting one sample with 1,142 genes) and intersected to 15,162 genes shared in all samples. The dataset was split into train, validation, and test sets with 35, 9, and 5 WSIs, respectively, keeping multiple slides of the same patient within a split. We used the 240 x 240 pixel image tiles centered around Visium spots as provided by \cite{Jaume2024HEST-1k:Analysis}, and log1p-transformed the gene expression counts. Image augmentation was applied to the training set; specifically, we resized image tiles to 224 x 224 pixels, applied a horizontal flip ($p=0.5$), 4-way 90 \degree rotation ($p=0.5$ each), random affine transformation (translate 0.02, scale 0.97 - 1.03, $p=1.0$), and color jittering (brightness 0.2, contrast 0.2, saturation 0.1, hue 0.05; p=0.5). We then applied ImageNet normalization with mean (0.485, 0.456, 0.406) and standard deviation (0.229, 0.224, 0.225). Validation and test data were only resized and normalized.\\
For external validation, we obtained n = 440 TCGA COAD samples, using the first WSI per sample. We used \texttt{lazyslide} \cite{Zheng2026LazySlide:Analysis} for detection of tissue regions and tiled to match the spot size of Visium data.\\
iCMS genes sets were retrieved from \cite{Joanito2022scRNA}, and stromal and immune gene sets were selected from Heiser et al. \cite{Heiser2023MolecularTumors}.

\subsection{Prediction architecture and training}
 To predict spatial gene expression, we employ UNI-2h \cite{Chen2022Self-SupervisedHistopathology}, which is based on ViT-g/14  as encoder, and combine it with per image tile gated attention pooling followed by a gene-specific prediction head (Fig. \ref{fig:model_overview}): 
 \begin{align*}
\text{UNI-2h ViT:}&\quad
  T = f_{\mathrm{enc}}(\mathrm{tile})
  && \text{H\&E tile $\to$ patch tokens $T$} \\
\text{Patch attention:}&\quad
  a = \mathrm{softmax}\!\big(w(\tanh(VT)\odot\sigma(UT))\big)
  && \text{weights $a$; $V,U,w$ learnable} \\
&\quad
  z = a^{\mathsf{T}} T
  && \text{pooled embedding $z$} \\
\text{Gene head:}&\quad
  \hat{y}_g = h_g(z)
  && \text{MLP $h_g$ $\to$ log1p of gene $g$}
\end{align*}
 Specifically, we pool ViT patch tokens on each input image tile with gated attention-based MIL pooling \cite{Tomczak2018Attention-basedLearning} using a single gating branch and a hidden dimension of 128, resulting in one $d=1536$ vector per tile. Next, we apply one MLP head per gene, consisting of Linear (1536$\rightarrow$256), ReLU, and Linear (256$\rightarrow$1) layers. We optimized a joint MSE–Pearson (MSE + (1 - Pearson $r$)) loss and use AdamW as optimizer with learning rates $1\cdot10^{-5}$, $5\cdot10^{-4}$, $1\cdot10^{-4}$, for encoder, attention-pooling, and gene heads, respectively. Additionally, we employ weight decay, with $10^{-3}$ for the encoder, $10^{-3}$ for both attention and gene heads, and 0 for all one-dimensional parameters (norms, biases). We shuffled tiles for training and used a batch size of 512. To account for training differences between genes, we conducted a two-phase training regime: first, we trained the last 10 encoder parameter groups and the MLP gene heads, then we kept the encoder frozen and fine-tuned the MLP gene heads. Both phases were trained for $30$ epochs with early stopping (patience $0.15$ $\times$ epochs).\\
 To categorize the predictions for spatial subtype classification, we employed an $argmax$ classifier on the GE signatures. Let $\hat{y}_{i,g}$ denote the predicted log1p expression of gene $g$ for tile $i$. For each signature $G_s$ with $S \in \{$stroma, immune, iCMS2 upregulated, iCMS2 downregulated, iCMS3 upregulated, iCMS3 downregulated$\}$,
the raw signature score is
\[
\mathrm{raw}_{i,s}
= \frac{1}{|G_s|}\sum_{g \in G_s} \hat{y}_{i,g}.
\]
Next, we determined calibration scalars on the train set $\mathcal{C}$ to balance gene expression magnitudes for each signature:
\[
\mu_s = \frac{1}{|\mathcal{C}|}\sum_{i \in \mathcal{C}} \mathrm{raw}_{i,s},
\]
resulting in normalized signatures per spot:
\[
\mathrm{norm}_{i,s} = \frac{\mathrm{raw}_{i,s}}{\mu_s}.
\]
We then hierarchically apply an $\arg\max$ classification with:\\
 $\mathcal{S}_{c}=\{$stroma, immune, iCMS2 upregulated, iCMS2 downregulated, iCMS3 upregulated, iCMS3 downregulated$\}$,\\
$s_i^\ast=\arg\max_{s\in\mathcal{S}_{c}}\mathrm{norm}_{i,s}$,
and a normal epithelial class 
with a safety margin $R=2$:
\begin{flalign*}
&\begin{aligned}
\mathcal{R}_i \equiv\;&
\bigl(\mathrm{norm}_{i,\mathrm{icms2down}}\ge R\,\mathrm{norm}_{i,\mathrm{icms3up}}
      \land \mathrm{norm}_{i,\mathrm{icms2down}}\ge \mathrm{norm}_{i,\mathrm{icms2up}}\bigr) \\
&\lor
\bigl(\mathrm{norm}_{i,\mathrm{icms3down}}\ge R\,\mathrm{norm}_{i,\mathrm{icms2up}}
      \land \mathrm{norm}_{i,\mathrm{icms3down}}\ge \mathrm{norm}_{i,\mathrm{icms3up}}\bigr).
\end{aligned}
&&
\end{flalign*}

and classify with:
\[
\text{class}_i=
\begin{cases}
s_i^\ast, & s_i^\ast\in\{\mathrm{stroma},\mathrm{immune}\},\\[4pt]
\mathrm{normal}, & \mathcal{R}_i,\\[4pt]
\arg\max\limits_{s\in\{\mathrm{icms2up},\mathrm{icms3up}\}}\mathrm{norm}_{i,s}
\;\mapsto\;\{\mathrm{i2},\mathrm{i3}\}, & \text{otherwise}.
\end{cases}
\]
We applied the same classifier for the measured GE by replacing
$\hat{y}_{i,g}$ with measured GE per tile $x_{i,g}$.

To investigate which genes can be well predicted, we calculated Moran's I using the \texttt{squidpy} implementation \cite{Palla2022Squidpy:Analysis}.

\subsection{Concept discovery using a relaxed archetypal sparse autoencoder}

To disentangle concepts of the ViT, we trained an RA-SAE \cite{Fel2025ArchetypalModels} on the embeddings produced by the forward pass of the fine-tuned UNI2-h encoder and the Patch Attention module of the HEST-1k train set. In particular, we used the activations after the attention pooling of the patch embeddings $d_{input}=1536$ as input with per-feature standardization, which was reverted for the reconstruction. RA-SAE aims to learn a dictionary $D$, which is constrained to a convex hull of the input data with relaxation $\delta$. To determine this convex hull, we used $n_C$ k-means centroids using the \texttt{faiss} implementation. We fixed the hidden dimension of the RA-SAE to $d_{hidden}=8000$, thereby guaranteeing an overcomplete dictionary size ($d_{input} \ll d_{hidden}$). On this dictionary, we masked TopK activations. 

\begin{align*}
\text{Embeddings as input:}&\quad
  z = \mathrm{AttnPool}\!\big(f_{\mathrm{enc}}(\mathrm{tile})\big)
  && \text{\parbox[t]{5.5cm}{\raggedright
       UNI-2h $+$ patch attention\\
       $z\in\mathbb{R}^{d_{\mathrm{input}}}$, $d_{\mathrm{input}}=1536$}} \\
\text{Standardize:}&\quad
  \tilde{z} = (z-\mu)./\sigma
  && \text{per-feature; undone at reconstruction} \\
\text{RA-SAE:}&\quad
  Z = \mathrm{TopK}\!\big(f_{\mathrm{enc}}^{\mathrm{SAE}}(\tilde{z});K\big)
  && \text{\parbox[t]{5.5cm}{\raggedright
       sparse activations $Z$\\
       hidden size $d_{\mathrm{hidden}}=8000$}} \\
\text{Dictionary:}&\quad
  D\in\mathrm{conv}(\{c_j\}_{j=1}^{n_C})_{\delta}
  && \text{\parbox[t]{5.5cm}{\raggedright
       $c_j$: $k$-means centroids\\
       $\delta$: relaxation}} \\
\text{Reconstruct:}&\quad
  \hat{z} = ZD + b
  && \text{decode; then un-standardize}
\end{align*}

To determine hyperparameters, we performed a grid search over TopK $K \in \{16,32\}$, $\delta \in \{1,25,35\}$ and $n_C \in \{8,16,32\} \times 10^3$ and used $\delta{=}25$, $K{=}32$, $n_C{=}8000$ as final parameters.

For training of the RA-SAE, we optimized $\mathcal{L}_{\mathrm{RA\text{-}SAE}} = \mathcal{L}_{\mathrm{MSE}} - \lambda \cdot \mathcal{L}_{\mathrm{reanim}}$ with $\lambda = 10^{-2}$.
The reanimation loss is defined as $\mathcal{L}_{\mathrm{reanim}} = \frac{1}{Bm} \sum_{ij} z_j^{(i)} \cdot \mathbf{1}[\text{concept } j \text{ inactive in batch}]$, where $z$ is the pre-TopK, pre-activation value. We used AdamW with $\mathrm{lr} = 10^{-3}$, weight decay $=0$, batch size $4096$, gradient clipping 1.0 and training for  $\leq 500$ epochs. We employed early stopping on the validation $\mathcal{L}_{\mathrm{MSE}}$ with patience 15.
We compared RA-SAE to TopK SAE with the same parameters as baseline, which was optimized using $\mathcal{L}_{\mathrm{MSE}}$ and trained using the same parameters.
To evaluate the performance of the RA-SAE, we used metrics described by \cite{Fel2025ArchetypalModels}. 

\subsection{Prototype derivation and concept visualization}
To explain the predictions on the HEST-1k test set and the TCGA COAD set, we employ the LRP $\gamma$-rule implemented in \texttt{zennit} \cite{Anders2021SoftwareViRelAy} with $\gamma_{conv}=0$ and $\gamma_{linear}=0.25$ to compute relevances in the MLP per gene head over per-signature aggregated outputs. The identity mapping between embedding and prediction head was then replaced by the trained RA-SAE, yielding relevance scores on the sparse latent concepts:
\begin{align*}
\text{GE prediction output:}&\quad
  R(\hat{y}_{G_s}) = \sum_{g\in G_s}\hat{y}_g
  && \text{\parbox[t]{5.5cm}{\raggedright
       $\hat{y}_g$: predicted gene $g$\\
       $G_s$: signature}} \\
\text{MLP per gene $\to$ RA-SAE concepts:}&\quad
  R(D) = \mathrm{LRP}^{\gamma}\!\big(R(\hat{y}_{G_s})\big)
  && \text{relevances on concepts $D$} \\
\text{$\ell_1$-normalize:}&\quad
  \tilde{R}(D)=R(D)/\|R(D)\|_{1}
  && \text{per-tile $\ell_1$-normalize} \\
\text{PCA:}&\quad
  R_{T}(D)=\mathrm{PCA}\!\big(\tilde{R}(D)\big)
  && \text{50 principal components} \\
\text{$k$NN:}&\quad
  G = k\mathrm{NN}(R_{T}(D))
  && \text{\parbox[t]{5.5cm}{\raggedright
       15 cosine neighbors\\
       graph for Leiden}} \\
\text{Clustering:}&\quad
  \{\mathcal{C}_k\} = \mathrm{Leiden}(G)
  && \text{\parbox[t]{5.5cm}{\raggedright
       Cluster tiles in\\
       concept-relevance space}} \\
\text{Prototypes:}&\quad
  i_k^\ast = \arg\max_{i\in\mathcal{C}_k}
  \cos\!\big(R(D)_i,\bar{R}_k\big)
  && \text{\parbox[t]{5.5cm}{\raggedright
       $\bar{R}_k$: centroid of $\mathcal{C}_k$\\
       tile $i_k^\ast$: prototype}} \\
\text{RelMax concepts:}&\quad
  \mathcal{C}^\ast = \mathrm{TopK}_c\!\Big(
    \max_k \lvert\bar{R}_{k,c}\rvert
  \Big)
  && \text{\parbox[t]{5.5cm}{\raggedright
       $\bar{R}_k$: mean of $\tilde{R}(D)$ on $\mathcal{C}_k$\\
       $K=\#\{\mathcal{C}_k\}$}} \\
\text{Isolate concept $c$:}&\quad
  R_c = R(D)\odot\mathbf{1}_c,\quad
  \forall c\in\mathcal{C}^\ast
  && \text{\parbox[t]{5.5cm}{\raggedright
       propagate relevance\\
       only for concept $c$}} \\
\text{RA-SAE concept $\to$ input image:}&\quad
  R_c(\mathrm{tile}) = \mathrm{LRP}^{\gamma}\!\big(R_c\big)
  && \text{\parbox[t]{5.5cm}{\raggedright
       propagate relevance\\
       through SAE encoder,\\
       patch attention\\
       and reformulated ViT.}}
\end{align*}
Notably, LRP can be computed in one backward pass and can be viewed as LRP-modified gradient $\times$ input.

To determine prototype tiles, we computed relevances per gene signature at the hidden layer of the RA-SAE. For normal epithelium, we averaged relevances from iCMS2- and iCMS3 downregulated signatures.

Next, we calculate the first 50 principal components of the $\ell_1$ normalized relevances on concepts, followed by a kNN graph (\# neighbors = $15$ with cosine distance), and UMAP (min\_dist = 0.5 for HEST-1k COAD and min\_dist = 0.25 for TCGA COAD). To detect clusters of tiles, we employ Leiden clustering at resolution 0.5, resulting in $K=\mathcal{C}_k$ concepts. Per cluster, the $8$ tiles with highest cosine similarity to the cluster centroid in concept-relevance space were used as prototypes. Concepts were selected by their
maximum absolute mass over Leiden centroids, retained the top $K=\mathcal{C}_k$ concepts. Rows and columns were then ordered by maximum absolute mass. To visualize an isolated concept $c$
($R_c=R(D)\odot\mathbf{1}_c$), we propagated relevance through the SAE encoder,
patch attention, and reformulated ViT onto the input tile
\cite{Achtibat2023CRP}.

To compute relevances within the ViT-specific components such as attention, LayerNorm and gated MLP, we used CP-LRP \cite{Ali2022XAIPropagation}, implemented in \texttt{lxt} \cite{Achtibat2024AttnLRP:Transformers} and adapted for the \texttt{timm} UNI-2h class by replacing \texttt{SDPA attention} with manual matrix multiplication attention (Fig. \ref{fig:hest_expl}C).

Heatmaps were smoothed with \texttt{scipy.ndimage.gaussian\_filter} ($\sigma=3$), thresholded at the
median of positive smoothed values, and cropped to the bounding box of the largest 8-connected component
(\texttt{scipy.ndimage.label}). For each concept we show the top six tiles by concept
mass.
To derive per-gene explanations, we backpropagated only relevances of the the respective gene and visualized its concepts together with prototypes of the signature it belongs to.

To compare relevance-based concepts with activation-based concepts, we applied the following steps:
\begin{align*}
\text{RA-SAE activations:}&\quad
  Z = f_{\mathrm{SAE}}(z)
  && \text{\parbox[t]{5.5cm}{\raggedright
       TopK activations\\
       $z \in\mathbb{R}^{1536}$, $Z\in\mathbb{R}^{8000}$}} \\
\intertext{\centering process activations like relevances}
\text{Patch-token map:}&\quad
  M_c(i) = f_{\mathrm{enc}}^{\mathrm{SAE}}(t_i)_{c},\quad
  \forall c\in\mathcal{C}^\ast
  && \text{\parbox[t]{5.5cm}{\raggedright
       $t_i\in\mathbb{R}^{1536}$: $i$-th patch token\\
       $P$: \# patch tokens ($P{=}16^2$)\\
       $M_c\in\mathbb{R}^{P}$: concept-$c$ map}}
\end{align*}
The visualization of the concept crop on the upsampled image was conducted as for the relevances.
As activations do not allow for direct associations with GE (or any outputs), we computed PCC of all concepts within prototypes of the respective signature with predicted GE, similar to \cite{PICASSO}. 

To keep computations feasible on the very large TCGA set, we subsample tiles. First, we sample tissue classes per sample, aiming for 512 tiles per patient with a floor threshold, keeping all tiles if a 5\% draw falls below 64. On these tiles, we store relevances. To reduce the size of the kNN graph, we then apply the cosine farthest-point sampling \cite{Eldar1997TheSampling}, where each newly acquired sample maximizes the distance to already selected ones, aiming for 64 instances per patient and class. We apply this both on relevances and activations, aiming to reduce highly dominant tissue patterns, and instead sampling on less prevalent tissues.

To compute the importance of activation- and relevance-based concepts we apply the definition by Fel et al. \cite{Fel2025IntoGeometry}:
\begin{equation*}
  I_c(Z)=\frac{1}{N}\sum_{i=1}^{N} Z_{i,c},
  \qquad
  I_c(R)=\frac{1}{N}\sum_{i=1}^{N}\lvert R_{i,c}\rvert
  \qquad c=1,\ldots,H,
\end{equation*}
both for the post-TopK activation matrix $Z$ and the gene-summed LRP matrix $\in\mathbb{R}^{N\times H}$ over all channels $c$ of the RA-SAE and all subsampled tiles $N$.

\subsection{Computational evaluation}
Gene expression predictions were evaluated on the test set with the Pearson Correlation Coefficient (PCC) per gene. Only genes with a PCC $\geq 0.3$ were kept for classification. The classification of iCMS on bulk-level (summed expression) was obtained using the \texttt{ICMS.SSC} package from Tsantoulis et al. \cite{Tsantoulis2025ACancer}. For outcome stratification, we obtained survival data for the TCGA cohort from Liu et al. \cite{Liu2018AnAnalytics}. Subsequently, we fitted Cox proportional hazard models to predict overall survival, correcting for tumor stage. We fitted only models with a sufficient number of events ($\geq 15 $) per estimated covariate and verified proportional hazard assumptions via Schoenfeld residuals as implemented in \texttt{survival cox.zph}.

\subsection{Histopathological Evaluation}
All H\&E images of both the HEST-1k COAD test set and a selection of the TCGA COAD were evaluated alongside predictions by a board-certified pathologist (P.B.). P.B. also evaluated concepts and prototypes of TCGA COAD.

\newpage
\section{Results}
\subsection{Spatial transcriptomics uncovers heterogeneous iCMS states in colorectal cancer}
\begin{figure}[h]
    \centering
    \includegraphics[width=\linewidth]{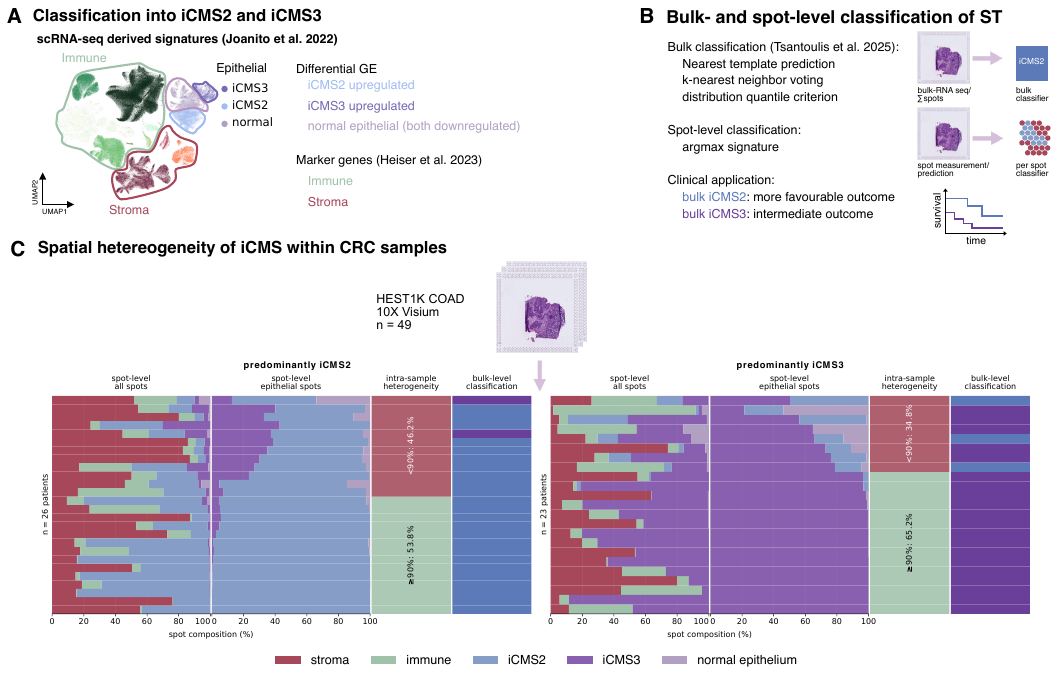}
    \caption{\textbf{A} scRNA-seq analysis by \cite{Joanito2022scRNA} revealed three states within epithelial cells in CRC, comprising malignant iCMS2, iCMS3, and normal cells, which are characterized by a corresponding set of GE signatures. To determine immune and stromal compartments, we retrieved marker genes from \cite{Heiser2023MolecularTumors}. \textbf{B}. Based on the malignant epithelial signatures, multiple classifiers for bulk RNA-seq have been developed \cite{Tsantoulis2025ACancer}, allowing for stratifying larger cohorts. To extend this classification to spot-level ST data, we developed a hierarchical $argmax$ classifier based on GE signature. iCMS2 has a more favourable outcome compared to iCMS3. \textbf{C} We reanalyzed ST data from \cite{Jaume2024HEST-1k:Analysis}, both on a spot-level and on aggregated measurements using the kNN nearest classifier, detecting highly prevalent intra-sample heterogeneity between both iCMS2 and iCMS3 on the spot-level, while the bulk-level mostly reflected the major population.}
    \label{fig:icms_intro}
\end{figure}

To assess the spatial heterogeneity of intrinsic CRC subtypes, we reanalyzed public spatial transcriptomics data from the HEST1-k COAD cohort (n = 49, \cite{Jaume2024HEST-1k:Analysis}) and applied GE-based classification both per spot and on pseudo-bulked measurements. Using the kNN classifier (with nearest class mode) revealed that approximately half of the samples were classified as iCMS2 and the other half as iCMS3 at the bulk level (Fig. \ref{fig:icms_intro}C). However, spot-level classification showed substantial heterogeneity within individual samples: among samples classified as iCMS2 at the bulk level, 46\% contained a mixture of both subtypes (<90\% of spots assigned to the same class, Fig. \ref{fig:icms_intro}C). For predominantly iCMS3 samples, we observed intra-sample heterogeneity for 35\% of all samples.

\subsection{ Concept-based explanations bridge local and global ViT interpretability}
\begin{figure}[h]
    \centering
    \includegraphics[width=1\linewidth]{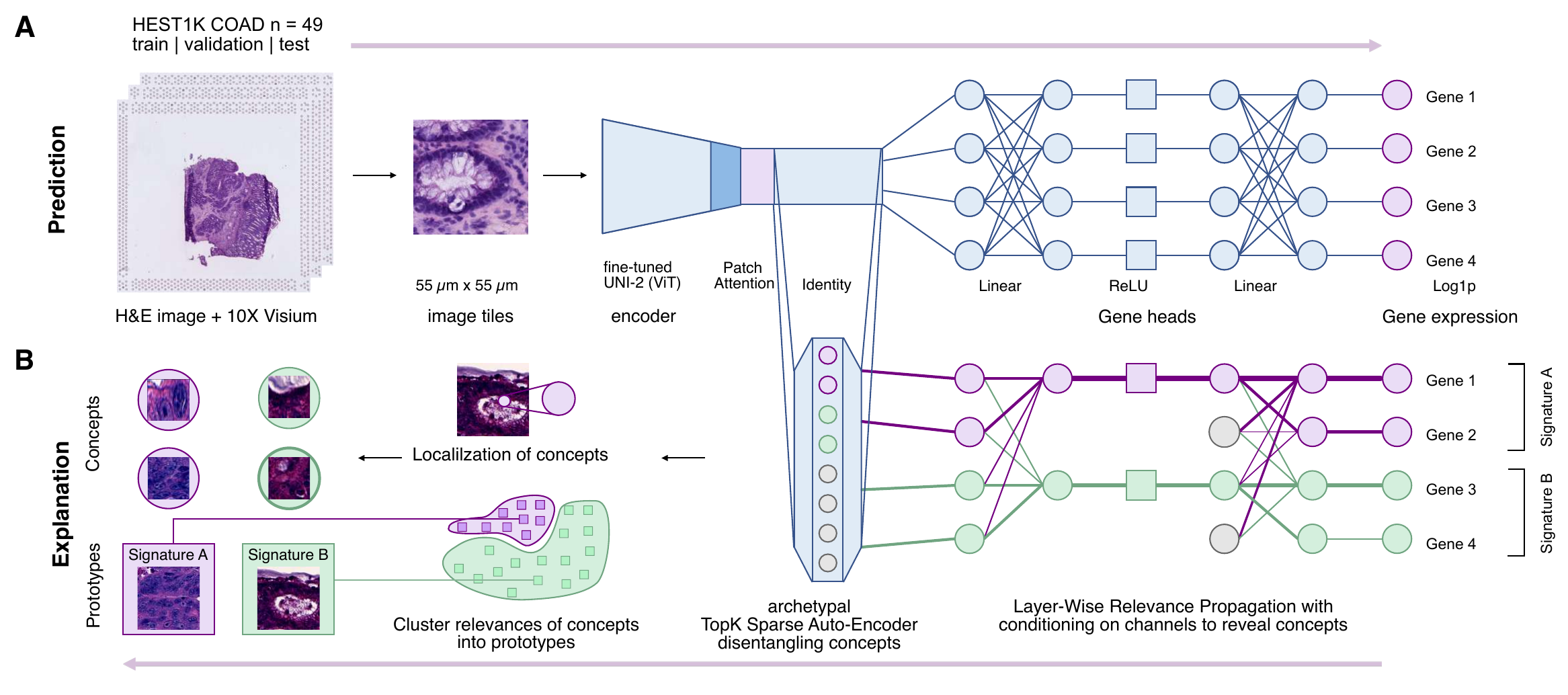}

        \caption{\textbf{A} For prediction, we split the 10X Visium HEST-1k COAD \cite{Jaume2024HEST-1k:Analysis} dataset into train, validation, and test sets. We developed a framework for GE prediction from H\&E tiles based on a fine-tuned UNI-2h \cite{Chen2022Self-SupervisedHistopathology} encoder, followed by patch attention \cite{Tomczak2018Attention-basedLearning}. Per-gene MLP heads predict stroma, immune, and iCMS signatures, preserving intra-signature variance.
\textbf{B} To explain the predictions, we apply LRP per signature \cite{Bach2015LRP}. As ViTs are encoding entangled concepts, we employ an RA-SAE \cite{Fel2025ArchetypalModels} to disentangle relevances into interpretable concepts. Next, we conduct Leiden clustering on the PCA-transformed relevances to derive monosemantic prototypes as global explanations. For these prototypes, we localize concepts in the input images.}
    \label{fig:model_overview}
\end{figure}
As virtual ST data is becoming more widely available, we developed a prediction framework (Fig. \ref{fig:model_overview}A) designed to provide both accurate predictions and meaningful explanations (Fig. \ref{fig:model_overview}B). Specifically, we fine-tuned a DINO-v2-based UNI-2h pathology FM for CRC and used patch attention and a per-gene linear head to predict log1p-transformed gene expression for the signatures shown in Fig. \ref{fig:icms_intro}A at the level of individual image tiles. The explanatory framework can be applied to other ViT-based architectures post-hoc. To disentangle concepts encoded in ViTs, we train an archetypal TopK sparse autoencoder (RA-SAE) \cite{Fel2025ArchetypalModels} on the spatial patch tokens processed with a patch attention module \cite{Tomczak2018Attention-basedLearning}. To link the latent representations of the RA-SAE to the prediction task, we apply layer-wise relevance propagation (LRP, \cite{Bach2015LRP}) from the predicted signatures back to the input images. This framework provides complementary global and local explanations: Leiden clustering of PCA-transformed concept relevances in the hidden layer of the RA-SAE yields representative prototype images without imposing prior assumptions, thereby revealing global model behavior. In turn, localizing individual concepts within the input image disentangles the individual concepts contributing to each prototype.

\subsection{Virtual transcriptomics recovers spatial gene expression patterns}
\begin{figure}[b!]
    \centering
    \includegraphics[width=1\linewidth]{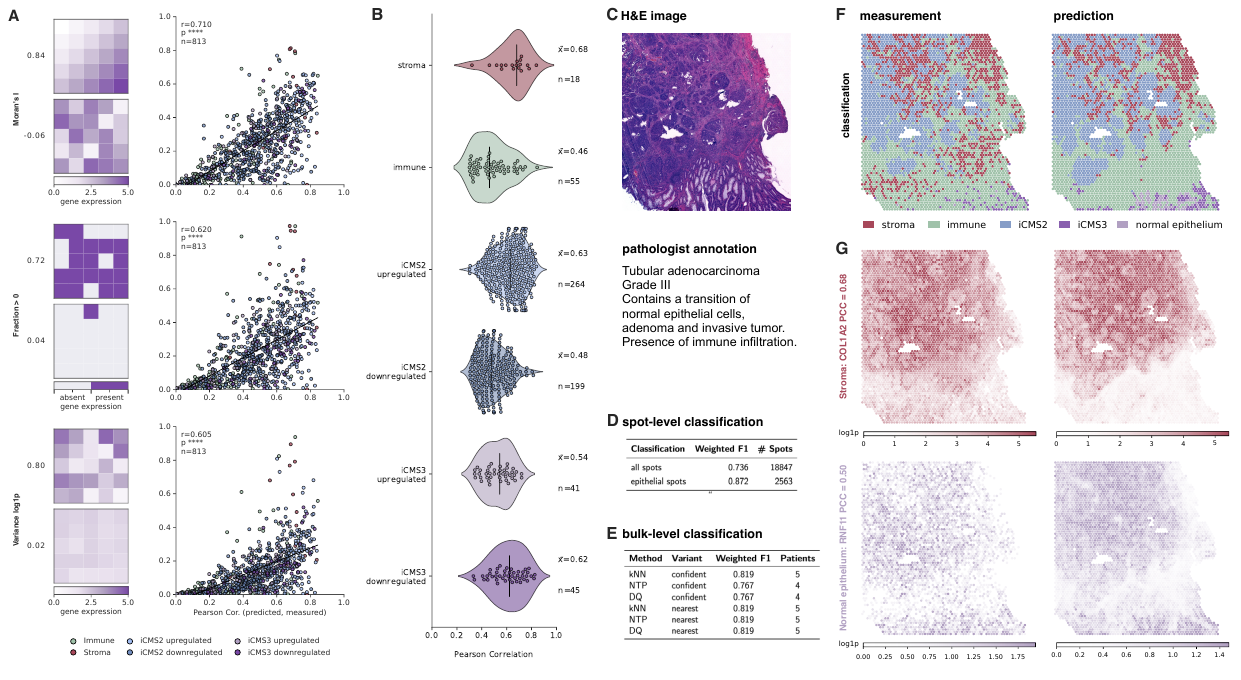}
    \caption{\textbf{Accurate prediction of GE and cell type identity is correlated with properties of spatial GE}. \textbf{A} Predictability is correlated with properties of ST data. To determine predictability, Pearson correlation between predicted
and measured GE was computed. This predictability is positively correlated with
spatial variability, the occurrence of non-zero GE, and GE
variance. \textbf{B} Our framework accurately predicts clinically relevant signatures, achieving median Pearson Correlations between measurements between 0.46 and 0.68 across different signatures, after filtering for genes with a PCC $\geq 0.3$. \textbf{C} Examplary H\&E image with pathologist annotation. \textbf{D} Comparing tissue identity revealed moderate weighted F1 scores for all cell types and high performance for the epithelial subset. \textbf{E} Bulk-level classification of aggregated spatial predictions was highly congruent with aggregated measurements. \textbf{F} Spatial overlap for hierarchical $argmax$ classification into tissue classes. \textbf{G} Predictions and measurements for representative genes showed relatively high spatial congruence.
}
    \label{fig:spatial_pred} 
\end{figure}

While recent work \cite{Nonchev2026DeepSpot-M:Histology} has demonstrated that large pathology FM predict transcriptome-wide GE from H\&E, prediction accuracy varies substantially across genes, and the determinants of gene predictability remain elusive. To focus on biologically interpretable transcriptional programs while reducing model complexity, we first selected gene features with known biological functions for prediction, resulting in 813 genes across six biological signatures (Fig. \ref{fig:icms_intro}). We then investigated which properties make genes predictable from histology. Prediction performance, measured by PCC on the test set, was positively associated with spatial autocorrelation (Moran's I), GE instance occurrence frequency, and GE variance (Fig. \ref{fig:spatial_pred}A). For downstream tasks, we retained only genes with PCC $\geq$ 0.3, resulting in 622 genes. Our model achieved state-of-the-art accuracy, with PCC values ranging from 0.46 to 0.68 across different signatures (Fig. \ref{fig:spatial_pred}B). To verify predictions, H\&E images were annotated by a pathologist (Fig. \ref{fig:spatial_pred}C, complete annotation of the HEST-1k COAD test set in SFig. \ref{fig:he_pred}). We then applied our hierarchical $argmax$ classifier to distinguish stromal and immune compartments and to classify epithelial spots as normal epithelium, iCMS2, or iCMS3. Comparison of measured and predicted ST data revealed moderate agreement for all tissue types and particularly strong concordance within the epithelial compartment (Fig. \ref{fig:spatial_pred}D). We further aggregated predictions at the sample level and performed iCMS classification based on work by \cite{Tsantoulis2025ACancer}, achieving high weighted F1 scores of 0.767-0.819 (Fig. \ref{fig:spatial_pred}E). Spatially, tissue-compartment assignments showed strong agreement between measured and predicted GE profiles (Fig. \ref{fig:spatial_pred}F, full test set in SFig. \ref{fig:he_pred}). Moreover, predicted expression maps of individual genes exhibited smooth spatial gradients that may reflect underlying expression patterns obscured by measurement sparsity (Fig. \ref{fig:spatial_pred}G).

\subsection{Virtual ST of TCGA COAD reveals iCMS heterogeneity associated with overall survival}
\begin{figure}[h!]
    \centering
    \includegraphics[width=\linewidth]{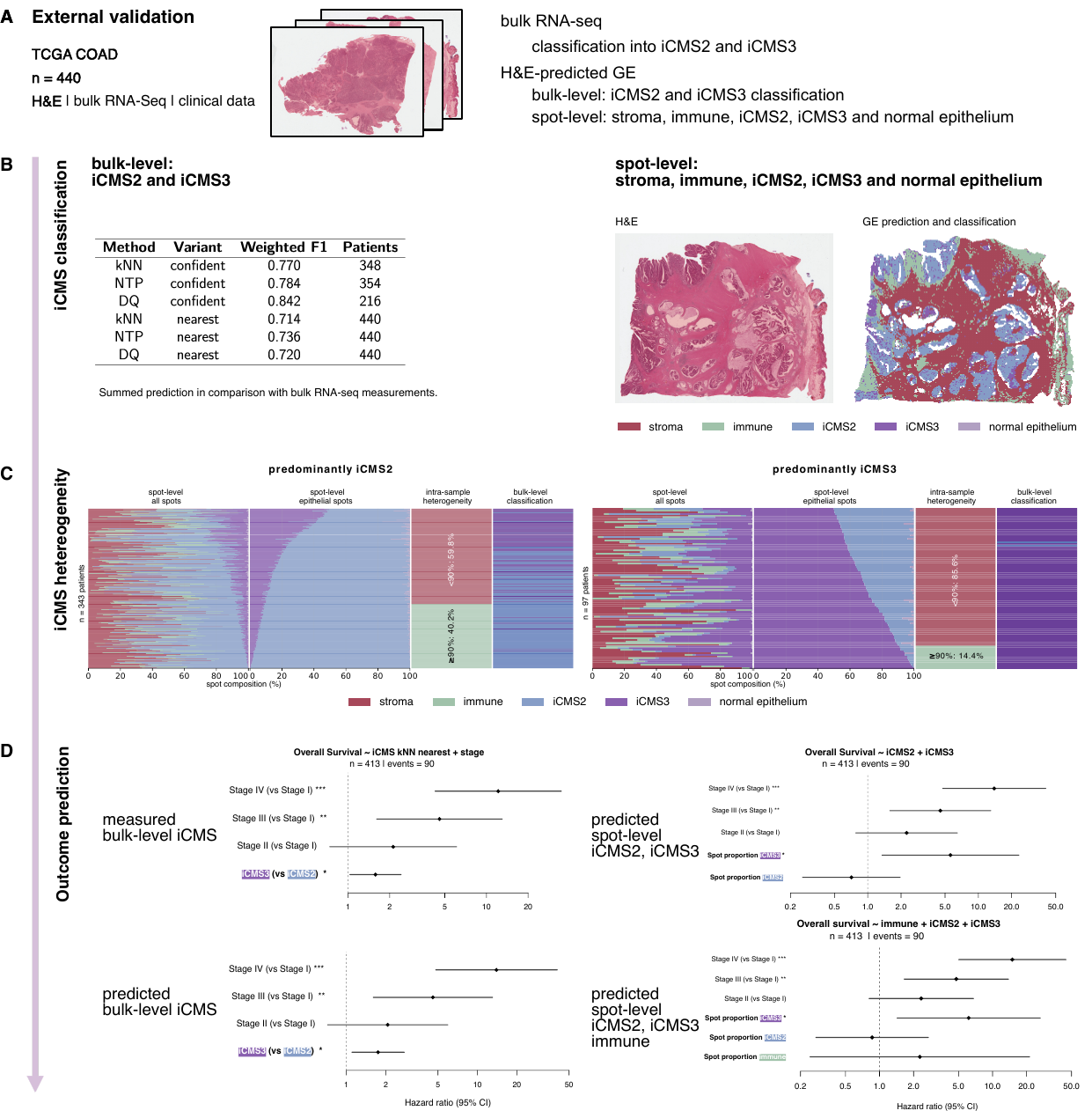}
    \caption{\textbf{Virtual ST on TCGA reveals iCMS heterogeneity and outcome stratification}. \textbf{A} For external validation, we retrieved H\&E, bulk RNA-seq, and clinical data of the TCGA COAD cohort ($n = 440$, \cite{Liu2018AnAnalytics}) and applied our H\&E prediction framework and subsequent classification. \textbf{B} Comparing predicted with measured bulk-level iCMS classification demonstrated relatively high weighted F1 scores. Predicted spot-level classification aligned well with H\&E-detectable tissue classes. \textbf{C} Our methods allows to deconvolute iCMS identity on a tissue basis, revealing iCMS heterogeneity that can partially reclassify bulk-level iCMS signals. Bulk-level displays aggregated predictions classified with kNN nearest mode. \textbf{D} Both measured and predicted bulk-level iCMS status can stratify overall survival. Interestingly, the proportion of iCMS3 spots was able to separate outcome as well, but the proportion of immune compartments could not stratify outcome. Stars indicate significance: *** - p-value $\leq 0.001$, ** - p-value $\leq 0.01$, * - p-value $\leq 0.05$.}
    \label{fig:tcga_val}
\end{figure}
To validate our model on unseen external data, we applied it to H\&E images of the TCGA COAD cohort (n = 440, Fig. \ref{fig:tcga_val}). We compared bulk-level iCMS classification derived from aggregated spot-level predictions with classifications based on bulk RNA-seq measurements and observed high precision across different classification strategies (Fig.\ref{fig:tcga_val}B). 

Spatially, spot-level classifications generated by our model corresponded to visually distinct regions in the H\&E images, enabling the construction of a virtual ST atlas of the TCGA cohort. Consistent with our observations in HEST-1k, many tumors exhibited substantial intra-tumoral heterogeneity, containing mixtures of iCMS2 and iCMS3 regions (Fig. \ref{fig:tcga_val}C). Although most samples were predominantly composed of iCMS2 spots ($n = 343$), a considerable fraction of these were nevertheless classified as iCMS3 at the bulk level. This discrepancy indicates that bulk-level iCMS classifiers capture not only the major spatial class, but likely encode an additional effect of class-specific GE expression magnitude. 

Previous studies have linked iCMS subtypes to clinical outcomes such as overall survival, prompting us to investigate whether our predictions retained this prognostic signal (Fig. \ref{fig:tcga_val}D). Indeed, Cox regression analyses yielded comparable significant hazard estimates for iCMS3 when applied to measured and predicted bulk-level expression profiles. Moreover, spot-level predictions revealed a dosage effect, with increasing fractions of iCMS3 regions associated with progressively higher hazard ratios.

\clearpage
\subsection{Relaxed archetypal TopK SAE extracts compact and interpretable morphological concepts}

Concept-based explanation methods remain largely unexplored in digital pathology. Most current models rely on ViTs as encoders, but attributing relevances directly to ViT encoder output neurons leads to highly spread-out relevances across all encoder dimensions. While clustering these relevances using PCX identifies visually coherent prototypes, the underlying concepts remain strongly entangled (SFig. \ref{fig:vit_encoder_pcx}).

Sparse TopK autoencoders have recently been proposed to disentangle such representations. However, Fel et al.\cite{Fel2025ArchetypalModels}  demonstrated that concept representations derived from activations (not relevances) are prone to initialization instability. They therefore proposed the relaxed archetypal SAE (RA-SAE), which constrains dictionary atoms to the convex hull of the encoder embeddings. To balance this constraint with reconstruction, their formulation includes a tunable relaxation parameter $\delta$.

We evaluated a large set of hyperparameters and found that stronger relaxation than reported by \cite{Fel2025ArchetypalModels} was necessary to achieve reconstruction performance comparable to that of TopK SAE (Table \ref{tab:rasae-grid-full}). Compared with TopK SAE, RA-SAE exhibited fewer dead codes and increased consistency, suggesting that the learned concepts remained closer to the underlying data manifold. Additionally, across four experiments, the best-reconstructing RA-SAE showed higher dictionary consistency. RA-SAE also learned concepts with lower intrinsic dimensionality, as measured by concept rank, indicating that individual concepts may correspond to more structured and potentially compositional regions of activation space.  However, dictionary coherence, which quantifies redundancy between dictionary atoms, was slightly increased in RA-SAE, consistent with the findings of Fel et al. \cite{Fel2025ArchetypalModels}.
The structure within SAE activations showed a slight decrease in connectivity, yielding less complex reconstructions, but activations that negatively interfered with each other were strongly decreased. For all subsequent experiments, we employed an RA-SAE with $\delta=25$, $k=32$, and $n_C = 8 \cdot 10^3$ k-means centroids, selected on the basis of reconstruction fidelity.

\subsection{A morphological concept atlas links gene expression phenotypes to histopathology}

\begin{figure}[b!]
    \includegraphics[width=1\linewidth]{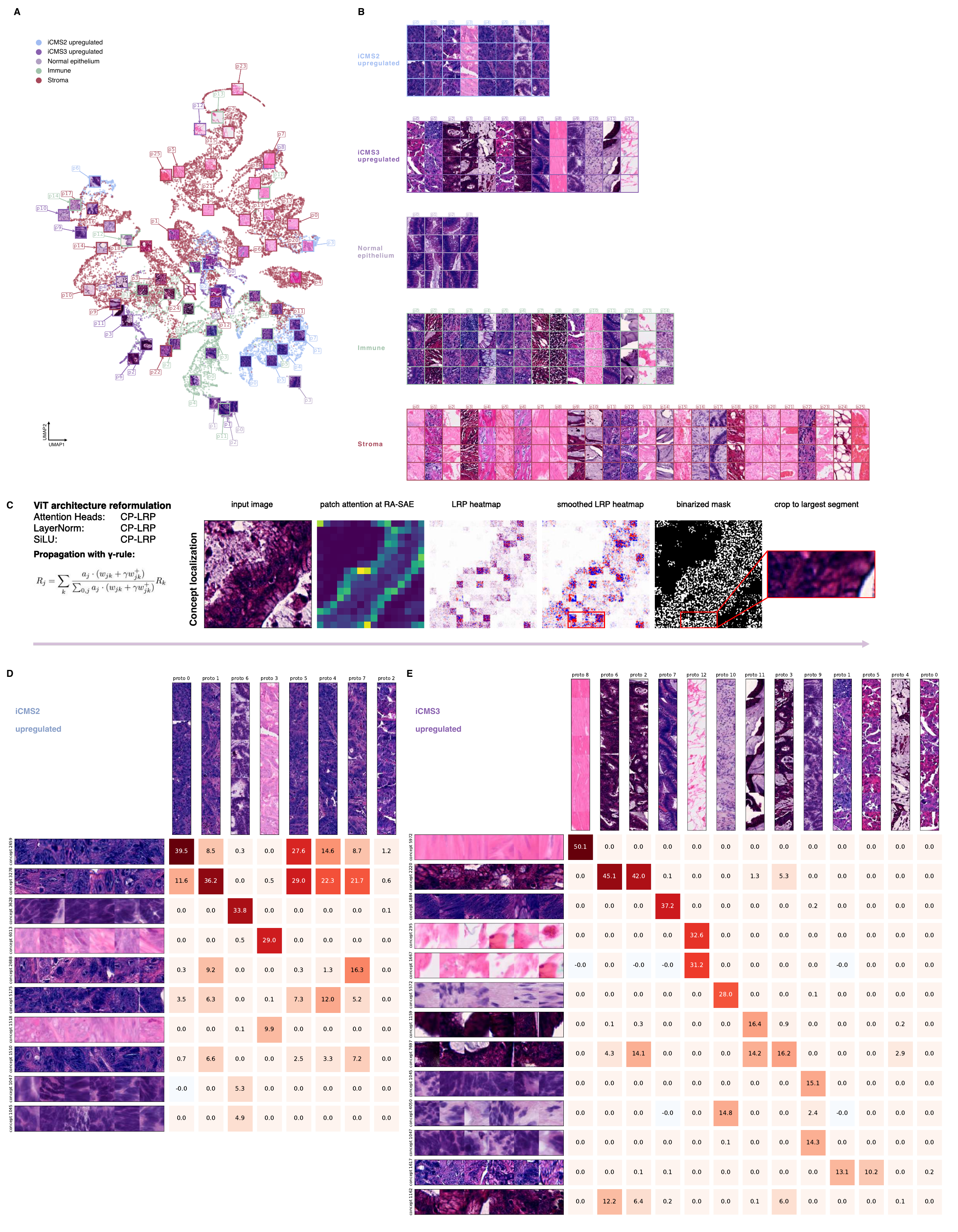} 
    %\addtocounter{figure}{-1}
\end{figure}
\begin{figure} [t!]
\caption{\textbf{HEST-1k COAD concept and prototype detection on RA-SAE relevances.} \textbf{A} UMAP embedding of relevances of all signatures of the HEST-1k COAD dataset at the hidden dimension of the RA-SAE. Prototype tiles are highlighted. \textbf{B} Prototypes per signature. \textbf{C} To localize concepts in images, we use CP-LRP to reformulate the ViT architecture, allowing for propagation using the Gamma rule. Additionally, we display patch attention weights at the RA-SAE. To obtain local explanations, we smooth the LRP heatmap and focus on the section with the largest positive attribution, revealing large nuclei in this example. \textbf{D-E} Concepts and prototypes for iCMS2 and iCMS3.
}
\label{fig:hest_expl}
\end{figure}

Linking molecular phenotypes to histopathological features is challenging. For iCMS, evidence for a clear association between subtype and morphology remains elusive. From a functional analysis of iCMS2/3 GE signatures, Tsantoulis et al. \cite{Tsantoulis2025ACancer} concluded that iCMS2 contains tubular adenoma pathway genes, hinting at a localisation at the crypt bottom, while iCMS3 is associated with the serrated sessile lesion pathway and gastric metaplasia, locating to the crypt top.

To uncover the relationship between molecular phenotypes and morphology, we first investigated the HEST-1k dataset by propagating gene-level across five signatures through our model. When tracing relevances from the MLP gene head to the SAE, 71\% got attributed to RA-SAE concepts, while the remainder was attributed to the RA-SAE error. Visualizing the resulting relevances at the RA-SAE concepts as a proxy for global model behaviour revealed a separation by molecularly confirmed tissue identity, albeit with partial overlap (Fig. \ref{fig:hest_expl}A). To identify representative prototypes for each tissue class, we clustered the samples within each class separately (SFig. \ref{fig:hest_a4}), yielding image patches that largely exhibited the expected morphological characteristics (Fig. \ref{fig:hest_expl}B). 

Given the limited resolution of spot-based ST, individual image patches frequently contain mixtures of different tissue classes and molecular phenotypes. To localize concepts, we reformulated the ViT architecture using CP-LRP \cite{Ali2022XAIPropagation}, enabling continuous relevance propagation with the LRP $\gamma$-rule (Fig. \ref{fig:hest_expl}C). Our framework also allows us to investigate patch-level attention weights in the RA-SAE representation as well as to propagate concept relevances back onto the input image. As LRP signals are noisy, we smoothed the heatmaps, binarized the resulting relevance maps, and cropped them to the largest connected component. In this way, global explanations based on RA-SAE prototypes can be linked to the spatial contribution of individual RA-SAE concepts in the input image. 

Interestingly, iCMS2 prototypes (Fig. \ref{fig:hest_expl}D) were typically explained by combinations of multiple concepts, whereas iCMS3 prototypes (Fig. \ref{fig:hest_expl}E) were often dominated by individual concepts. This observation suggests that iCMS2 morphology may be more distributed across multiple histological patterns, while iCMS3 exhibits more distinctive visual hallmarks. However, many prototypes were strongly patient-specific (SFig. \ref{fig:hest_a4}), likely reflecting both the pronounced histopathological diversity across patients and the limited size of the HEST-1k cohort (SFig. \ref{fig:he_pred}).

\begin{figure}
    \includegraphics[width=1\linewidth]{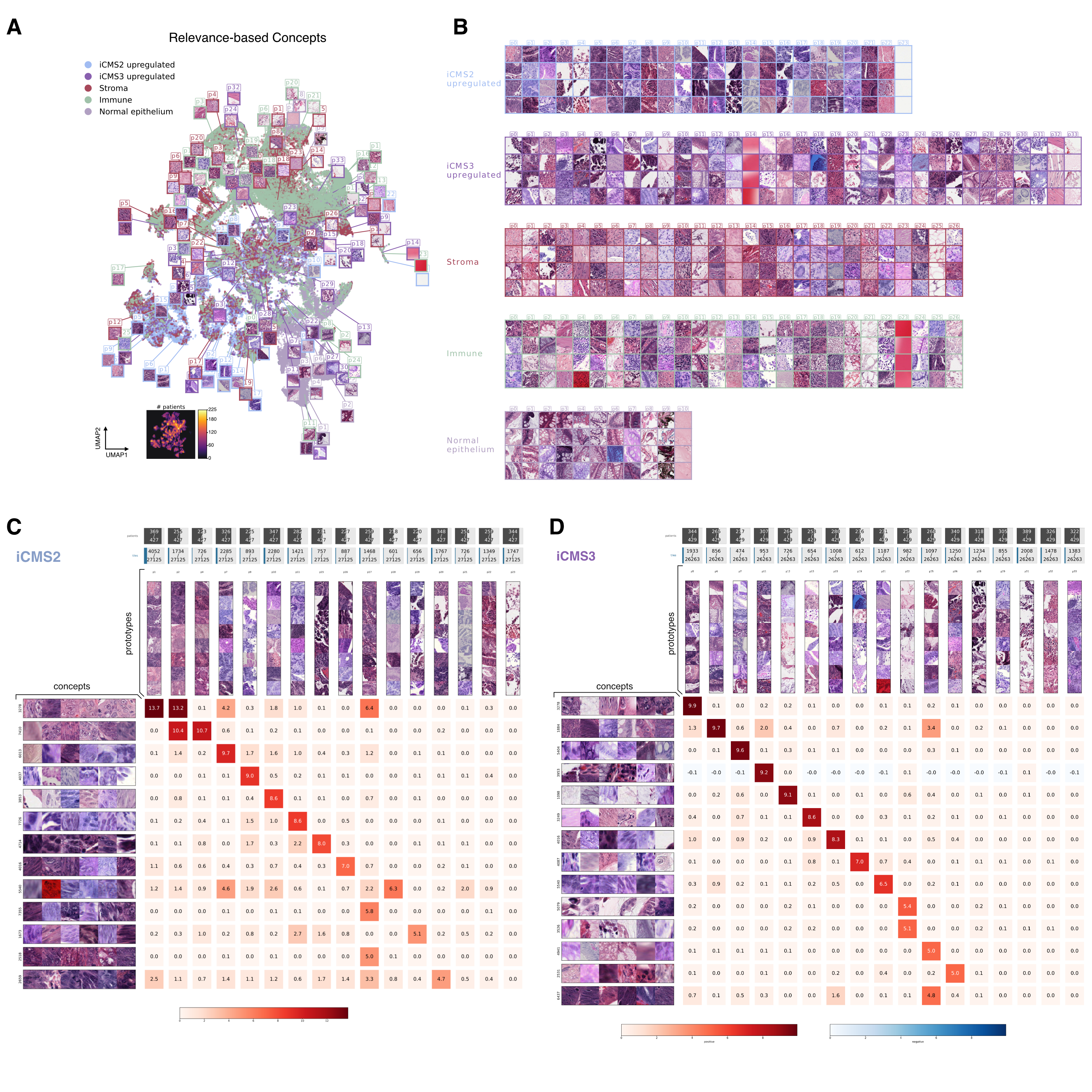}
    \caption{\textbf{TCGA COAD prototype detection on RA-SAE relevances.} \textbf{A} UMAP embedding of relevances of all signatures of the virtual ST TCGA COAD dataset at the hidden dimension of the RA-SAE. Prototype tiles are highlighted and a density map of the same embedding highlights high coverage by multiple patients. \textbf{B} Prototypes per signature. \textbf{C} iCMS2 and \textbf{D} iCMS3 prototypes and concepts, and the fraction of patients contributing to the prototype are displayed.}
\label{fig:tcga_expl}
\end{figure}

To overcome the limitations of small-scale ST datasets, we constructed a concept atlas for iCMS2/3, stromal, immune, and normal epithelial signatures on the TCGA COAD cohort, with representative prototypes for each signature. Relevances in the RA-SAE latent space showed intermediate separation in UMAP projection, indicating that individual image tiles often contain mixtures of molecular phenotypes (Fig. \ref{fig:tcga_expl}A). We additionally identified a small cluster of artefacts, including empty tiles resulting from failed tissue detection (p23 for iCMS2) as well as color artefacts (p14, p23 for iCMS3 and immune, Fig. \ref{fig:tcga_expl}B). 
Importantly, prototypes displayed substantial variation in staining patterns, demonstrating generalization capability and robustness to color variation.

To investigate the morphological correlates for iCMS2 and iCMS3, we restricted our analysis to prototypes that were present in $\geq 50\%$ of all patients and occurred in $\geq1\%$ of tiles within the corresponding tissue class; we applied a permissive threshold because each prototype covers only a small fraction of tiles (all prototypes are shown in SFig. \ref{fig:tcga_a4}). As some concepts could only be interpreted in the context of the surrounding tissue architecture, we analyzed concepts jointly with their associated prototypes. 

For iCMS2 concepts, we identified classical hallmarks of malignant epithelium, including enlarged nuclei and prominent nucleoli (p1, p2, p7, p8, p17),  as well as tubular growth patterns (p2, p4). Additional concepts captured various nuclei morphologies, including hyperchromatic, elongated and pseudostratified nuclei (p7). hyperchromatic and polygonal nuclei (p11) as well as a combination of elongated nuclei and small nucleoli (p16). Other concepts captured meso-scale morphologies, such as tumor cells without surrounding stroma (p9), intraluminal necrotic debris (p10), and artifacts such as dark-stained regions (p15).

Several concepts were shared between iCMS2 and iCMS3, including enlarged nuclei with prominent nucleoli (iCMS3 p5), elongated nuclei and small nucleoli (iCMS2 p16, i3 p25). However, iCMS3 also exhibited distinctive concepts that were absent from iCMS2, such as normal epithelium with transition to adenoma containing goblet cells (p6, p7, p25), larger areas with necrotic debris devoid of crypts (p19), as well as mucus with interspersed tumor and immune cells (p22, p28). In addition, we identified a prototype representing mixed tumor-stromal regions containing eosinophil granulocytes (p11).

Stromal tissue prototypes predominantly contained collagen fiber structures (p3, p4, p16, p18, p20), as well as connective tissue (p20, p21, p23, (SFig. \ref{fig:tcga_pcx_st_im_norm}A). Immune prototypes were overwhelmingly dominated by immune cells, but some also contained other stromal and epithelial components. Interestingly, while most immune concepts highlighted immune infiltrates, a small number explicitly focused on the morphology of the tumor- or goblet cells (SFig. \ref{fig:tcga_pcx_st_im_norm}B). Normal epithelial prototypes exhibited an intact crypt architecture (p2, p4) and regularly spaced enterocytes (p5), spanning a range of crypt and lumen sizes (SFig. \ref{fig:tcga_pcx_st_im_norm}C).

Together, our framework demonstrates that concept-based explanations of foundation models reveal rich associations between gene-expression programs and histopathological organization at scale. Importantly, many of these associations emerge without manual annotation or prior knowledge. Explanatory analyses of foundation models can thus uncover previously unrecognized links between molecular states and tissue morphology, while also providing insight into the visual features and internal representations underlying foundation model outputs.

\clearpage
\subsection{Can Concepts be better defined by relevance or activations?}

\begin{figure}[b!]
    \includegraphics[width=1\linewidth]{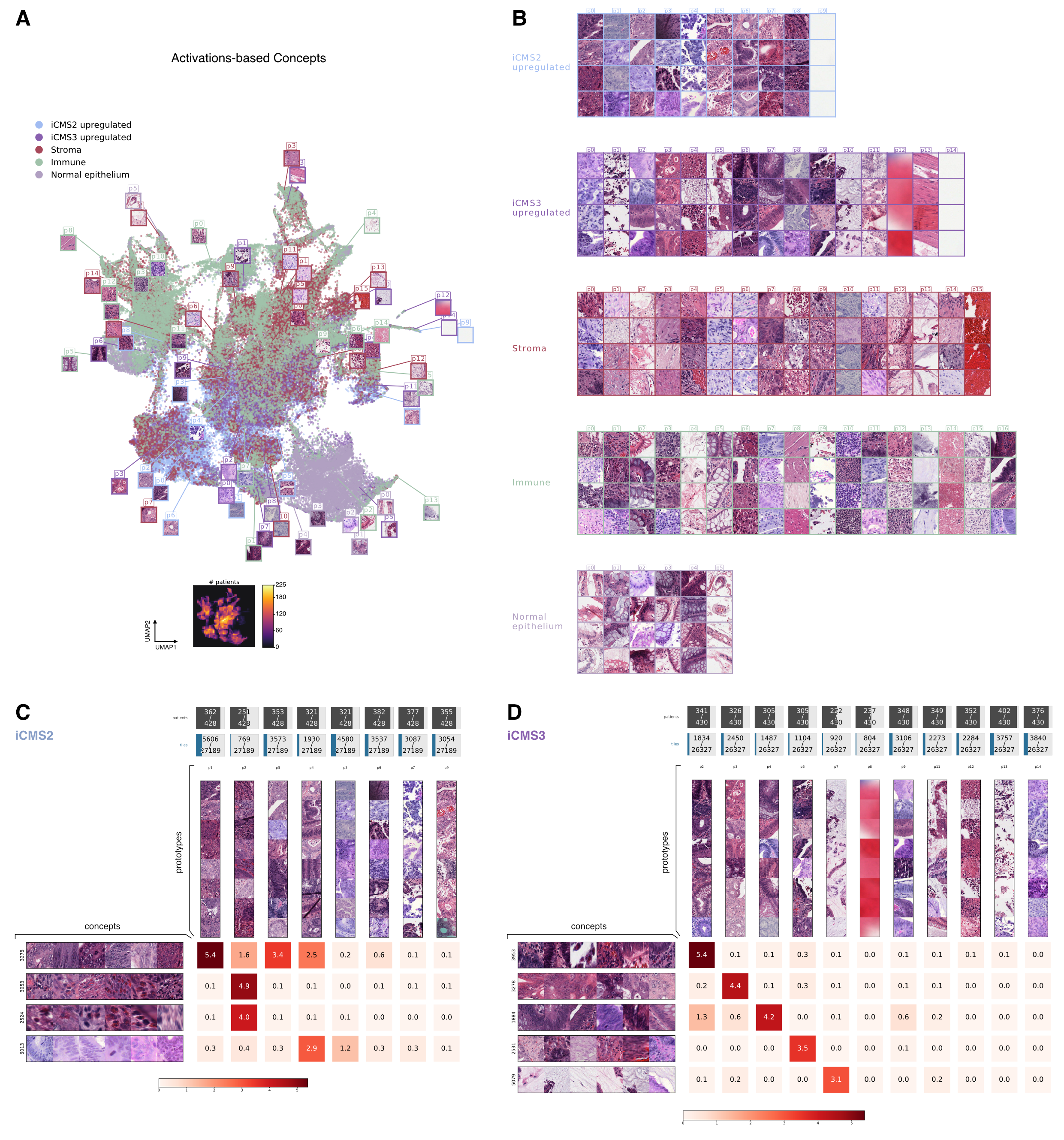} 
\caption{\textbf{Activation-based concepts in TCGA-COAD.} \textbf{A} UMAP embedding of activations of all signatures of the virtual ST TCGA COAD dataset at the hidden dimension of the RA-SAE. Prototype tiles are highlighted and a density map of the same embedding highlights high coverage by multiple patients. \textbf{B} Prototypes per signature. \textbf{C} iCMS2 and \textbf{D} iCMS3 prototypes and concepts, and the fraction of patients contributing to the prototype are displayed.}

\label{fig:tcga_umap_act}
\end{figure}

Existing work on DINOv2 ImageNet \cite{Fel2025IntoGeometry} and pathology FM concepts \cite{PICASSO} has defined concepts through latent activations rather than relevance scores. To compare our relevance-based framework with this line of work, we repeated the concept-discovery pipeline on TCGA COAD using RA-SAE activations instead of relevances. UMAP embeddings of activation-based concepts showed moderate separation between gene-expression signatures (Fig. \ref{fig:tcga_umap_act}A), similar to relevance-based concepts. However, prototypes detected at the same clustering resolution were less diverse (Fig. \ref{fig:tcga_umap_act}B), and concepts selected using the criteria of Section 3.6 yielded weaker morphological distinctions between iCMS2 and iCMS3. In addition, activation-based concepts contained a prominent cluster dominated by image artifacts (red blurry image, Fig. \ref{fig:tcga_umap_act}D). Nevertheless, some biologically meaningful concepts, such as a mucin-associated iCMS3 concept, were recovered by both approaches.

\begin{figure}[b!]
    \includegraphics[width=1\linewidth]{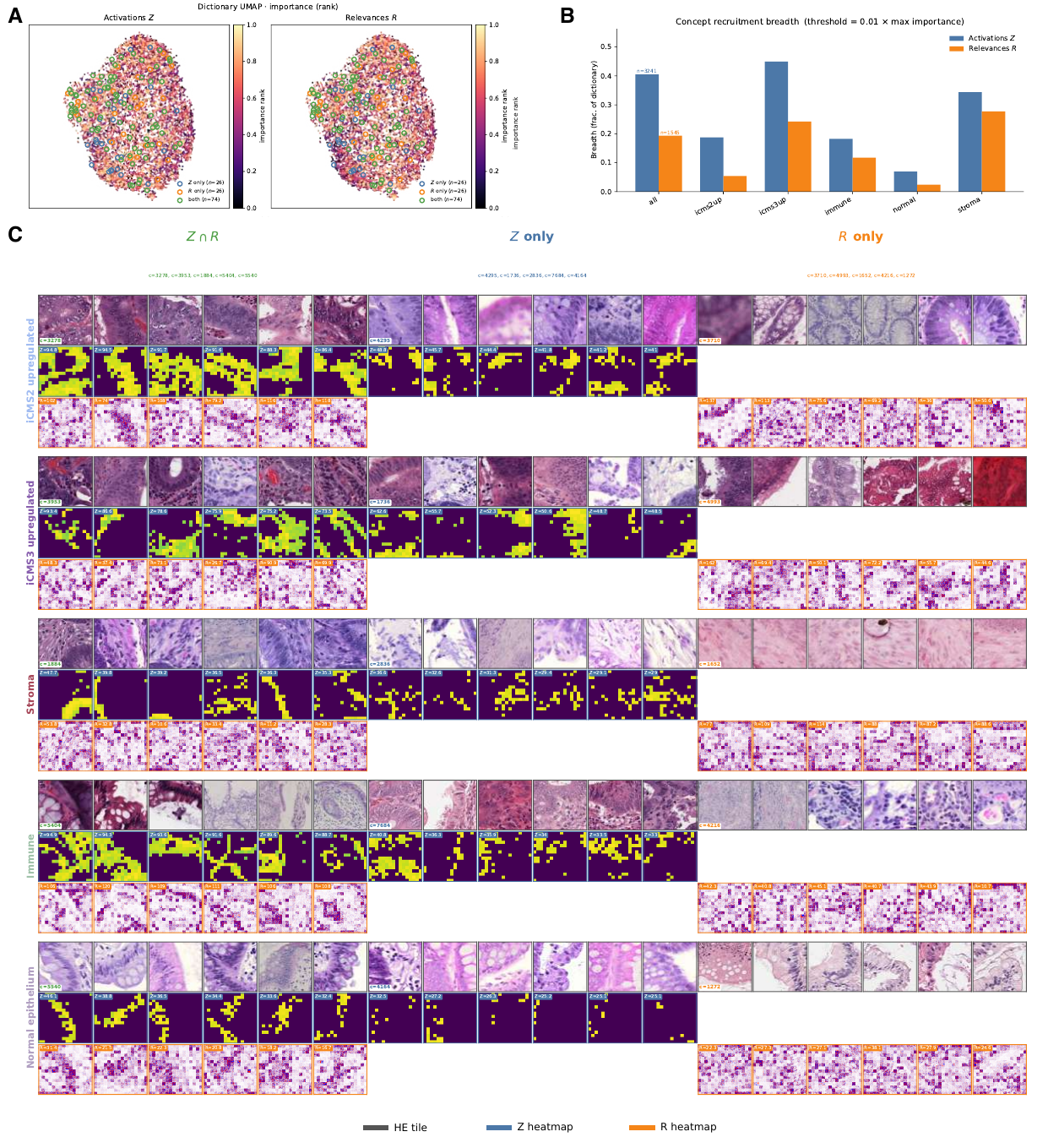} 
\caption{\textbf{Difference between RA-SAE concept usage by relevances and activations}. \textbf{A} UMAP on dictionary, concept overlap between topk with k=100 selected concepts. Colored by normalized importance rank for Z and R. Both are highly correlated regarding their importance rank (Spearman = 0.942) and overlap moderately (Jaccard index for the top-100 = 0.587). \textbf{B} Concept usage breadth for R and Z by signature. \textbf{C} Joint and disjoint concepts between relevance and activations. We selected from the top-100 concepts for each tissue class concepts with maximal importance and visualized them on tiles selected by max Z. Thereby, we investigate \#tissue classes $\times$ \{joint, Z only, R only\} different concepts.}
\label{fig:rel_act_tiles}
\end{figure}

Given the qualitative differences between relevance- and activation-based concepts, we analyzed their usage of the RA-SAE concept dictionary as previously described \cite{Fel2025IntoGeometry}. Ranking concepts by their importance revealed strong agreement between relevances (R) and activations (Z), with high correlation (Spearman correlation = 0.942) and a moderate overlap among the top 100 concepts (Jaccard index = 0.587, Fig. \ref{fig:rel_act_tiles}A). We next investigated whether the fraction of concepts utilized per signature differed between R and Z, and found consistently lower fractions of concepts utilized by R (Fig. \ref{fig:rel_act_tiles}B). This observation aligns with the intuition that the RA-SAE dictionary is learned from activations, whereas only a subset of these concepts is attributed relevance under the LRP decomposition \cite{Bach2015LRP}.

Several metrics have been proposed to compare concept usage in SAEs based on Z \cite{Fel2025ArchetypalModels}, and we extended them to R to further characterize these differences. R-based concepts exhibited lower negative interference and more refined internal structure, as measured by maximum cosine similarity and effective rank, although other metrics showed comparable or less favorable performance (Tab. \ref{tab:act-vs-rel}).

\begin{table}[h]
\centering
\caption{Activation ($Z$) vs.\ relevance ($R$) on the RA-SAE dictionary of all TCGA COAD tiles ($N{=}80176$, $H{=}8000$, top-$K{=}100$), summing relevances of all outputs.
Agreement: top-$K$ Jaccard $0.587$, Spearman importance $0.942$ Pearson importance $0.887$.
}
\label{tab:act-vs-rel}
\resizebox{\textwidth}{!}{%
\begin{NiceTabular}{c|rr|rrrr|rrrr}
\hline
 & \multicolumn{2}{c|}{\textbf{Structure within Z and R codes}} & \multicolumn{4}{c|}{\textbf{Concept usage}} & \multicolumn{4}{c}{\textbf{Structure within Z and R activated Concepts}} \\
\cline{2-11}
 & \textbf{Conn.\ $\uparrow$} & \textbf{Neg.\ Int.\ $\downarrow$} & \textbf{Imp.\ $\ell_1$} & \textbf{Imp.\ max} & \textbf{Breadth} & \textbf{$n_{\mathrm{act}}$} & \textbf{Intra $|\cos D|$ $\uparrow$} & \textbf{Max $|\cos|$ $\uparrow$} & \textbf{St.\ Rank $\downarrow$} & \textbf{Eff.\ Rank $\downarrow$} \\\hline
$Z$ & 0.745 & \underline{$1.10{\times}10^{6}$} & 619.0 & 4.331 & 0.405 & 3241 & \textbf{0.091} & \underline{0.576} & \textbf{11.83} & \underline{63.87} \\
$R$ & 0.745 & {\boldmath $2.60{\times}10^{5}$} & 310.1 & 4.567 & 0.193 & 1545 & \underline{0.090} & \textbf{0.655} & \underline{13.40} & \textbf{62.04} \\
\hline
\end{NiceTabular}%
}
\end{table}

Despite the substantial overlap in usage between R and Z, we next explored disjoint concept usage. This question is particularly interesting in light of the \textit{elsewhere} concepts empirically described by Fel et al. \cite{Fel2025IntoGeometry}, in which activations arise in patch tokens spatially separated from the object encoded by the Z-based concept. To assess whether R-based concepts exhibit similar behavior, we analyzed concepts that were either shared between Z and R or unique to one of the two representations (Fig. \ref{fig:rel_act_tiles}C). For Z-based concepts, we highlighted the firing patch tokens, while for relevance-based concepts, we attributed them to input images as described in Fig. \ref{fig:hest_expl}C. Notably, activation signals highlighted larger image regions compared to relevance signals. 

The shared concepts showed substantial spatial overlap between the R and Z signals in the input images, although the R maps appeared to be noisier. The Z-only concepts for iCMS2, iCMS3, and immune signatures aligned well with visually apparent morphological features. However, we also observed Z-specific concepts that lacked an obvious morphological interpretation. For instance, in normal epithelium, a concept shared between Z and R highlighted hyperchromatic epithelial cells. The presented Z-only concept for normal epithelium did not map to a clear morphological pattern, whereas the R-only concept seemed to capture lumen structures.
While these examples suggest qualitative differences between Z- and R-based concepts, a systematic comparison will require quantitative metrics. 

Contrary to previous approaches based on Z \cite{PICASSO}, our framework enables concepts to be linked directly to GE predictions. We therefore explored how individual genes contribute to concepts, for both Z and R. Following the approach of Kim et al. \cite{PICASSO}, we first computed the correlation between Z-based concepts and predicted expression of \textit{MYC}, a hallmark gene of iCMS2. Visualizing Z-based concepts cropped to patch tokens together with iCMS2 prototypes revealed several concepts moderately correlated with predicted GE, which corresponded to cells with visible nuclei in various tissue contexts (Fig. \ref{fig:gene_act_rel}A) and partially overlapped with the signature-wide concepts (Fig. \ref{fig:tcga_umap_act}C). Repeating this for R-based concepts yielded a greater number of correlated concepts reflecting more diverse morphological structures (Fig. \ref{fig:gene_act_rel}). Importantly, correlations with predicted \textit{MYC} GE remained low for the majority of all Z- and R-based concepts, indicating that our approach identifies gene-specific concepts in either scenario (Fig. \ref{fig:gene_act_rel}B). Overall, while Z- and R-based concepts rely on largely overlapping concept dictionaries, R-based concepts appear to yield more diverse and biologically meaningful prototypes, facilitating the interpretation of downstream predictions.

\clearpage
\section{Discussion}

This work extends prototypical concept-based explanations \cite{Achtibat2023CRP} and prototype discovery \cite{Dreyer2024PCX} to ViT-based spatial prediction models and establishes a general strategy for post hoc, local and global explanation of spatially distributed predictions. While concept-based interpretability approaches have shown promise in image classification settings, their application to transformer architectures and spatial molecular prediction tasks has remained largely unexplored. By adapting relevance propagation to a ViT-based gene expression prediction framework and combining it with sparse autoencoder-based concept discovery, we demonstrate that latent representations can be linked to interpretable tissue concepts without constraining the predictive model during training. Importantly, concepts and prototypes emerge from the learned representation space post hoc, preserving model flexibility and predictive performance. In contrast to recent concept-discovery approaches that operate independently of downstream tasks \cite{PICASSO}, our framework explicitly conditions concept discovery on model predictions through relevance maximization \cite{Achtibat2023CRP}. Consequently, the concepts identified by our approach are derived using a ViT-based pathology foundation model and are also adapted to specific gene signatures, thus revealing which concepts mainly drive particular predictions through clustering in explanation space. Furthermore, per-instance attribution maps provide spatially resolved concept localizations, linking global concept structure to local morphological evidence.

Using colorectal cancer as a use case, we demonstrate that the proposed framework accurately predicts clinically relevant iCMS, stromal, and immune signatures from H\&E images and recovers spatial patterns of iCMS heterogeneity, as validated in a large external cohort. Moreover, concept-based explanations identify morphologically coherent concepts associated with predicted transcriptional programs. Expert pathologist review showed that these concepts capture not only recognizable histomorphological structures, including glandular architecture, mucinous regions, stromal compartments, and areas characterized by nuclear atypia, but also distinct cellular content such as infiltrating immune cells. Finally, our relevance-based concept atlas lays the groundwork for systematic future studies on the link between morphological structures and molecular phenotypes.

Our work sits at the intersection of three rapidly evolving fields: digital pathology, spatial biology, and concept-based explainability. Recent studies have demonstrated increasingly accurate prediction of spatial GE from routine histology, while foundation models have substantially improved representation learning and generalization across pathology tasks. In parallel, interpretability research has progressed from pixel-level attribution towards concept-centric approaches, including concept relevance propagation and sparse autoencoder-based decomposition of latent representations in large vision models \cite{Achtibat2023CRP,Fel2025IntoGeometry}. However, these developments have remained largely disconnected. Virtual spatial transcriptomics methods have primarily focused on predictive performance, whereas concept discovery has been explored predominantly in natural-image domains. To our knowledge, this study is among the first to combine transformer-based virtual spatial transcriptomics with concept-level explanation. By doing so, it extends interpretability beyond single-image classification to spatially distributed biological signals, where tissue context and intra-sample heterogeneity are intrinsic aspects of the prediction task.

Several limitations warrant consideration. First, virtual spatial transcriptomics remains fundamentally constrained by the quality, scale, and diversity of available training data, introducing uncertainty into both model training and evaluation. Although the HEST-1k COAD cohort contains a broad spectrum of colorectal cancer morphologies, including varying degrees of differentiation and distinct growth patterns, the number of samples available for model development remains modest. Moreover, there is currently no computationally efficient strategy for deriving an optimal set of prototypes and concepts, limiting the comparability of R- and Z-based concepts. Alternative clustering approaches, resampling-based stability analyses, and recently proposed concept-quality metrics may improve the reproducibility and interpretability of discovered concepts in future studies \cite{Gao2020SelectiveClustering, Gu2021Cola:Framework, Parisini2026LeakageModels}. 

Multiple methodological directions could further strengthen the proposed framework. First, our current implementation aggregates relevance across genes within transcriptional signatures, potentially obscuring gene-specific explanatory patterns. Future approaches could jointly model genes, spatial locations, and RA-SAE hidden dimensions to uncover additional biological structure and disentangle the contributions of individual genes to spatial molecular phenotypes. Second, improvements in relevance propagation may further enhance the fidelity of local explanations; in particular, the structure of the top layers of our architecture is compatible with reference-value based attribution \cite{Letzgus2021TowardModels}. Finally, we observed that relaxed archetypal sparse autoencoders required substantially weaker archetypality constraints than previously reported to achieve satisfactory reconstruction quality in pathology representations, motivating future work on regularization strategies tailored to biomedical image embeddings. More generally, concept-based explanation methods still lack universally accepted measures of faithfulness \cite{Kumar2025MeasuringExplanations}. Establishing rigorous evaluation frameworks for discovered concepts remains an important open challenge and will be essential for the reliable deployment of concept-based explanations in biomedical applications.

More broadly, a theoretical understanding of the geometric and topological relationships between pathology images, activations, GE signatures, and relevances remains a desideratum. Natural images are thought to lie on low-dimensional manifolds \cite{IanGoodfellowandYoshuaBengioandAaronCourvilleDeepLearning}, and gradient-based explanations have been hypothesized to lie in their tangent spaces \cite{Bordt2022TheExplanations}. The authors suggest estimating such manifolds using autoencoders, although this may introduce architectural priors. A joint theory of the different data modalities and model properties is of particular interest, as the inference of biological features from H\&E images reverses the causal chain through which genetic, transcriptomic, and proteomic alterations give rise to morphological features. However, explanations obtained by backpropagating predicted molecular measurements to the input images follow the computational graph in the opposite direction and may therefore provide insight into how molecular programs manifest morphologically. Furthermore, CRP induces conditional graphs in the explanation space, while biological systems are themselves organized as networks spanning multiple molecular modalities. Bridging explanation graphs and biological networks may therefore provide a promising route towards multimodal and causal models of tissue organization, linking concept-based explanations to biological processes across different molecular layers.

In conclusion, we present a framework for concept-based explanation of transformer-based spatial prediction models and demonstrate its utility for virtual spatial transcriptomics. By combining transformer-aware relevance propagation with sparse autoencoder-based concept discovery, the framework provides both local and global insights into the tissue morphologies associated with predicted molecular programs. Beyond colorectal cancer, the approach is broadly applicable to spatial prediction tasks in computational pathology and may serve as a foundation for the interpretable analysis of emerging pathology foundation models and multimodal architectures.

\section{Authorship statement}
A.M. and J.T. conceived the study with input from R.A. and M.D. under the supervision of T.W., D.H., S.L., W.S. and T.G.K.. J.T. and A.M. implemented the experiments. J.T. implemented the training pipeline. C.F. and H.P. conducted preceding experiments. P.B. verified the classification into tissue classes and annotated H\&E images. The manuscript was written by A.M., T.G.K. and J.T. with input from all authors. This work is based on the results and code of the Master's thesis of J.T..
\section{Funding}
A.M. and T.G.K. were supported by the Federal Ministry of Education and Research and AM by the Deutsche Forschungsgemeinschaft (DFG, German Research Foundation) funded Research Training Group (RTG) 2424/CompCancer - project number 377984878. 
\section{Acknowledgments}
We would like to thank Dagmar Kainmüller and Nils Blüthgen for their supervision and fruitful discussions. 
We also thank the HPC of the Berlin Institute of Health for computing time.
\newpage
\noindent\section{Supplementary Materials}
\renewcommand{\thepage}{S\arabic{page}}
\renewcommand{\thesection}{S\arabic{section}}
\renewcommand{\thetable}{S\arabic{table}}
\renewcommand{\thefigure}{S\arabic{figure}}
\setcounter{figure}{0}
\renewcommand{\figurename}{Supplemental Figure}

\begin{table}[h]
\centering
\caption{RA-SAE hyperparameter grid search on UNI-2h patch attention-pooled embeddings
vs.\ TopK baseline. All combinations of $K\in\{16,32\}$, $\delta\in\{1,25,35\}$,
$n_C\in\{8,16,32\}{\times}10^3$. Across all runs, best and
second best per metric are marked; the best R$^2$ among
RA-SAE runs is additionally marked by a framed row. One diverged run
($^\dagger$) is excluded from ranking. The hidden dimension was fixed at $d=8000$. Z - SAE
activations. OOD - out-of-distribution score. Stability (Hungarian aligned average cosine similarity) and maximum cosine similarity were computed for dictionaries of four experiments.}
\label{tab:rasae-grid-full}
\resizebox{\textwidth}{!}{%
\begin{NiceTabular}{c|rrr|rr|rrr|rrr|rr}
\hline
 & \multicolumn{3}{c|}{\textbf{Hyperparameters}} & \multicolumn{2}{c|}{\textbf{Sparse Reconstruction}} & \multicolumn{3}{c|}{\textbf{Consistency}} & \multicolumn{3}{c|}{\textbf{Structure within Concepts}} & \multicolumn{2}{c}{\textbf{Structure within $Z$}} \\
\cline{2-14}
 & \textbf{$\delta$} & \textbf{$K$} & \textbf{$n_C\,(10^3)$} & \textbf{R$^2$ $\uparrow$} & \textbf{Dead Z $\downarrow$} & \textbf{OOD $\downarrow$} & \textbf{Stability\ $\uparrow$} & \textbf{Max Cos.\ $\uparrow$} & \textbf{St.\ Rank $\downarrow$} & \textbf{Eff.\ Rank $\downarrow$} & \textbf{Coh.\ $\downarrow$} & \textbf{Conn.\ $\uparrow$} & \textbf{Neg.\ Int.\ $\downarrow$} \\\hline
\multirow{2}{*}{\rotatebox{90}{\textbf{TopK}}} & -- & 16 & -- & 0.777 & 0.582 & 0.756 & -- & -- & 62.58 & 1277.5 & \underline{0.796} & \textbf{0.991} & 293.63 \\
 & -- & 32 & -- & \textbf{0.821} & 0.310 & 0.744 & 0.333 & 0.374 & 77.04 & 1288.9 & \textbf{0.729} & 0.960 & 789.70 \\
\hline
\multirow{18}{*}{\rotatebox{90}{\textbf{RA-SAE}}} & 1 & 16 & 8 & 0.607 & 0.281 & 0.507 & -- & -- & 2.06 & 303.2 & 0.997 & \underline{0.979} & 8.20 \\
 & 1 & 16 & 16 & 0.527 & 0.214 & 0.492 & -- & -- & \underline{1.96} & 279.9 & 0.994 & 0.970 & 14.74 \\
 & 1 & 16 & 32 & 0.371 & 0.148 & 0.497 & -- & -- & \textbf{1.77} & \textbf{270.8} & 0.997 & 0.968 & 12.66 \\
 & 25 & 16 & 8 & 0.727 & 0.063 & 0.608 & -- & -- & 13.62 & 1038.6 & 0.970 & 0.972 & \underline{7.61} \\
 & 25 & 16 & 16 & 0.729 & 0.034 & 0.615 & -- & -- & 14.47 & 1042.9 & 0.909 & 0.970 & \textbf{7.23} \\
 & 25 & 16 & 32 & 0.693 & 0.031 & 0.581 & -- & -- & 7.33 & 967.9 & 0.921 & 0.967 & 15.25 \\
 & 35 & 16 & 8 & 0.702 & 0.088 & 0.590 & -- & -- & 9.96 & 981.7 & 0.979 & 0.969 & 11.10 \\
 & 35 & 16 & 16 & 0.716 & 0.033 & 0.610 & -- & -- & 14.11 & 1008.6 & 0.917 & 0.968 & 9.50 \\
 & 35 & 16 & 32 & 0.704 & 0.029 & 0.582 & -- & -- & 7.91 & 959.6 & 0.896 & 0.966 & 13.20 \\
\cline{2-14}
 & 1 & 32 & 8 & 0.620 & 0.097 & 0.501 & -- & -- & 2.11 & 334.3 & 0.987 & 0.909 & 13.35 \\
 & 1 & 32 & 16 & 0.585 & 0.038 & \underline{0.483} & -- & -- & 2.53 & 304.6 & 0.994 & 0.899 & 16.90 \\
 & 1 & 32 & 32 & 0.587 & 0.019 & \textbf{0.463} & -- & -- & 2.34 & \underline{271.8} & 0.995 & 0.903 & 16.00 \\
 & 25 & 32 & 8 & \underline{0.780} & 0.008 & 0.643 & 0.533 & 0.580 & 19.84 & 1076.7 & 0.940 & 0.891 & 10.85 \\
 & 25 & 32 & 16 & 0.724 & 0.010 & 0.565 & -- & -- & 6.66 & 946.0 & 0.917 & 0.874 & 31.32 \\
 & 25 & 32 & 32 & $-1.9{\times}10^{6}\,^\dagger$ & 0.366 & 0.433 & -- & -- & 1.58 & 17.9 & 1.000 & 0.985 & 757135.94 \\
 & 35 & 32 & 8 & 0.760 & 0.013 & 0.622 & -- & -- & 12.53 & 1028.5 & 0.939 & 0.884 & 17.61 \\
 & 35 & 32 & 16 & 0.764 & \underline{0.005} & 0.615 & -- & -- & 15.76 & 1008.1 & 0.844 & 0.881 & 19.97 \\
 & 35 & 32 & 32 & 0.727 & \textbf{0.006} & 0.581 & -- & -- & 8.19 & 941.1 & 0.916 & 0.876 & 13.41 \\
\hline
\CodeAfter
  \begin{tikzpicture}
\draw [black, line width=1pt, rounded corners=2pt]
      (17-|2) rectangle (18-|15) ;
  \end{tikzpicture}
\end{NiceTabular}%
}
\end{table}

\begin{figure} [b!]
    \centering
    \includegraphics[width=\linewidth]{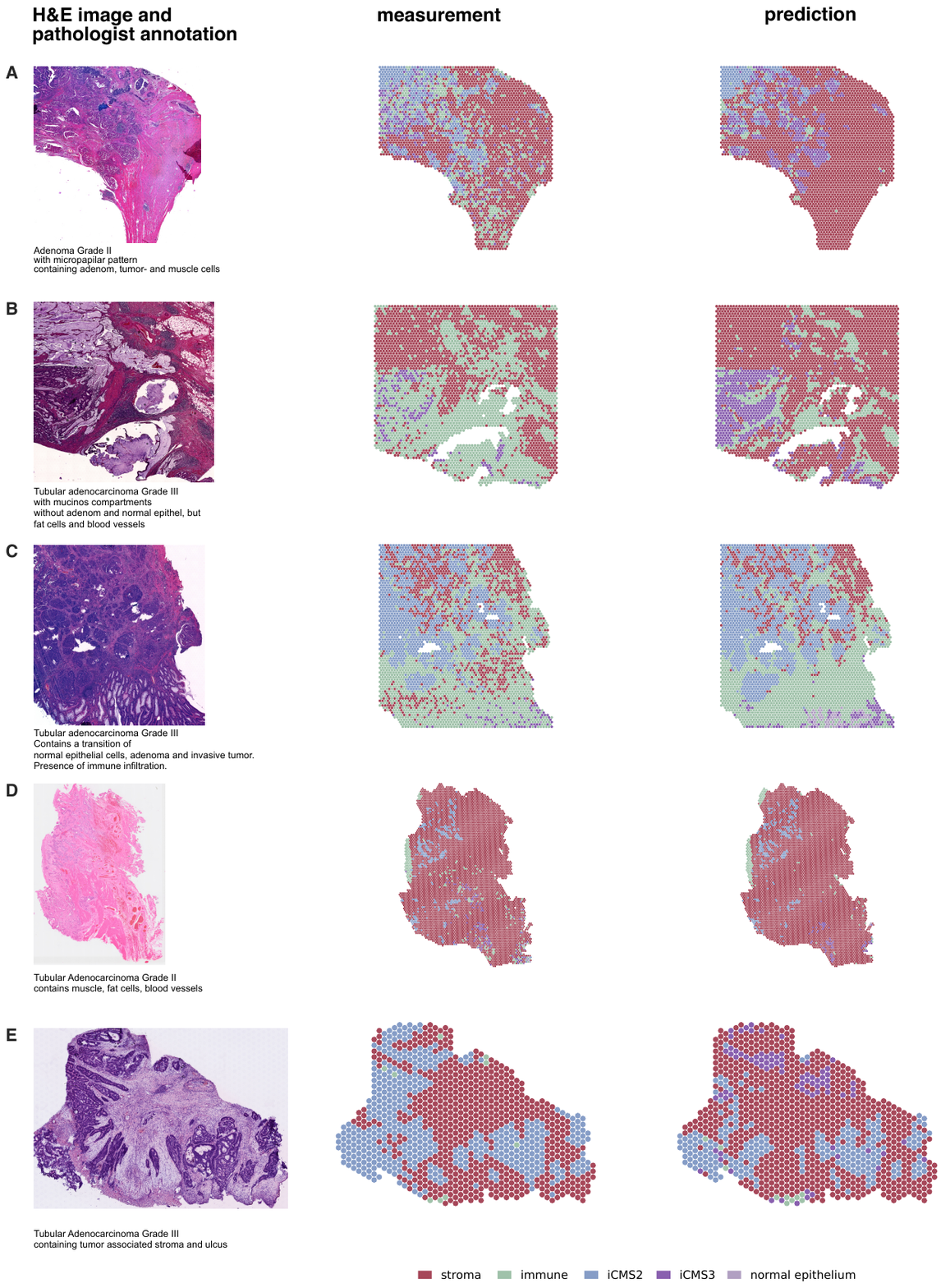}

\end{figure}
%\addtocounter{figure}{-1}
\begin{figure} [t!]
    \caption{
\textbf{Predictions align well with H\&E images}\\For all patients in the test set, we annotated H\&E images by a pathologist, revealing high heterogeneity across patients. Using a simple \texttt{argmax} classifier, we classifiy all spots based on GE signatures into broad tissue classes. 
}
    \label{fig:he_pred}
\end{figure}
\begin{figure}
    \centering
    \includegraphics[width=0.5\linewidth]{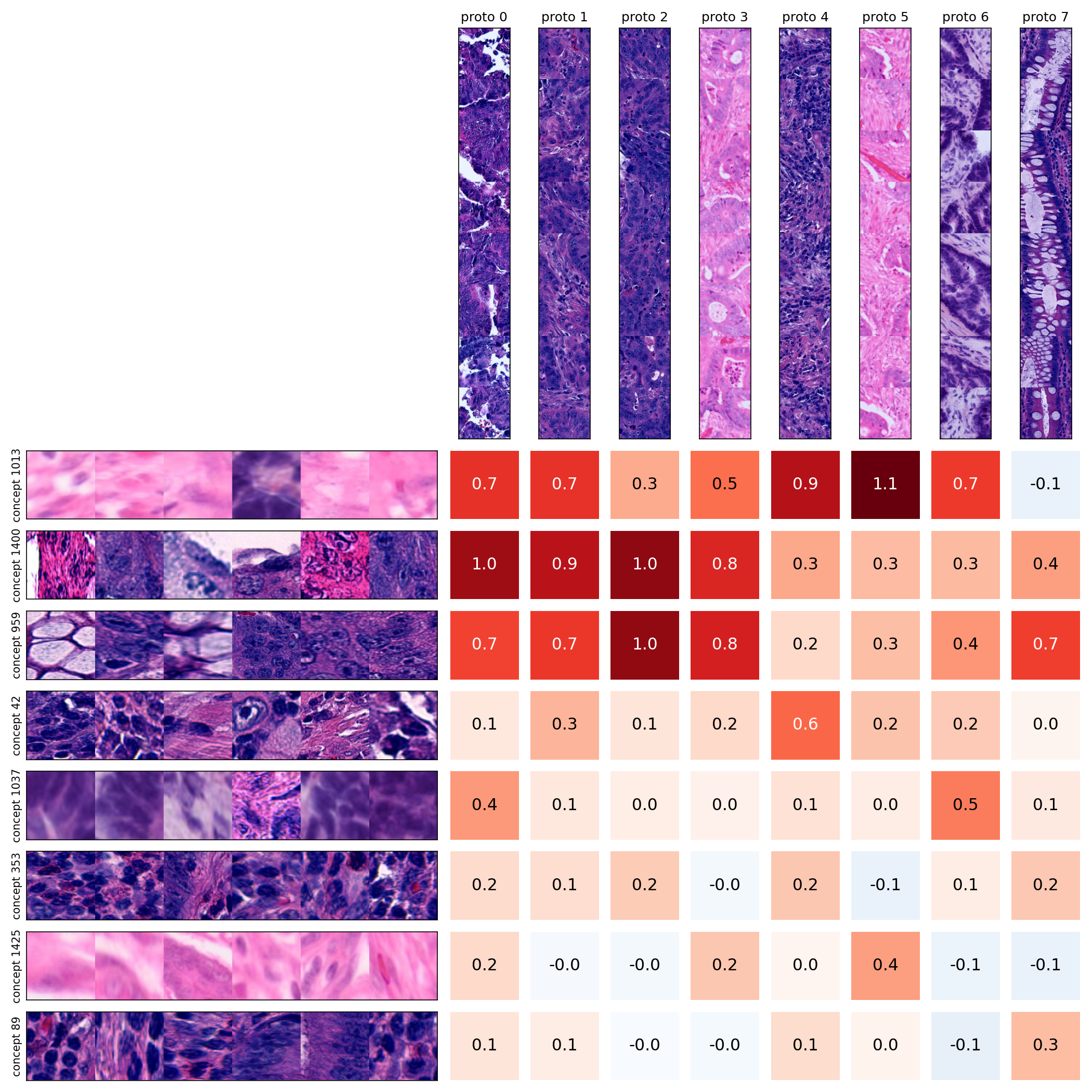}
    \caption{Entangled concepts in ViT encoder output. Example from iCMS2 upregulated signature.}
    \label{fig:vit_encoder_pcx}
\end{figure}

\begin{figure}
    \centering
    \includegraphics[width=0.9\linewidth]{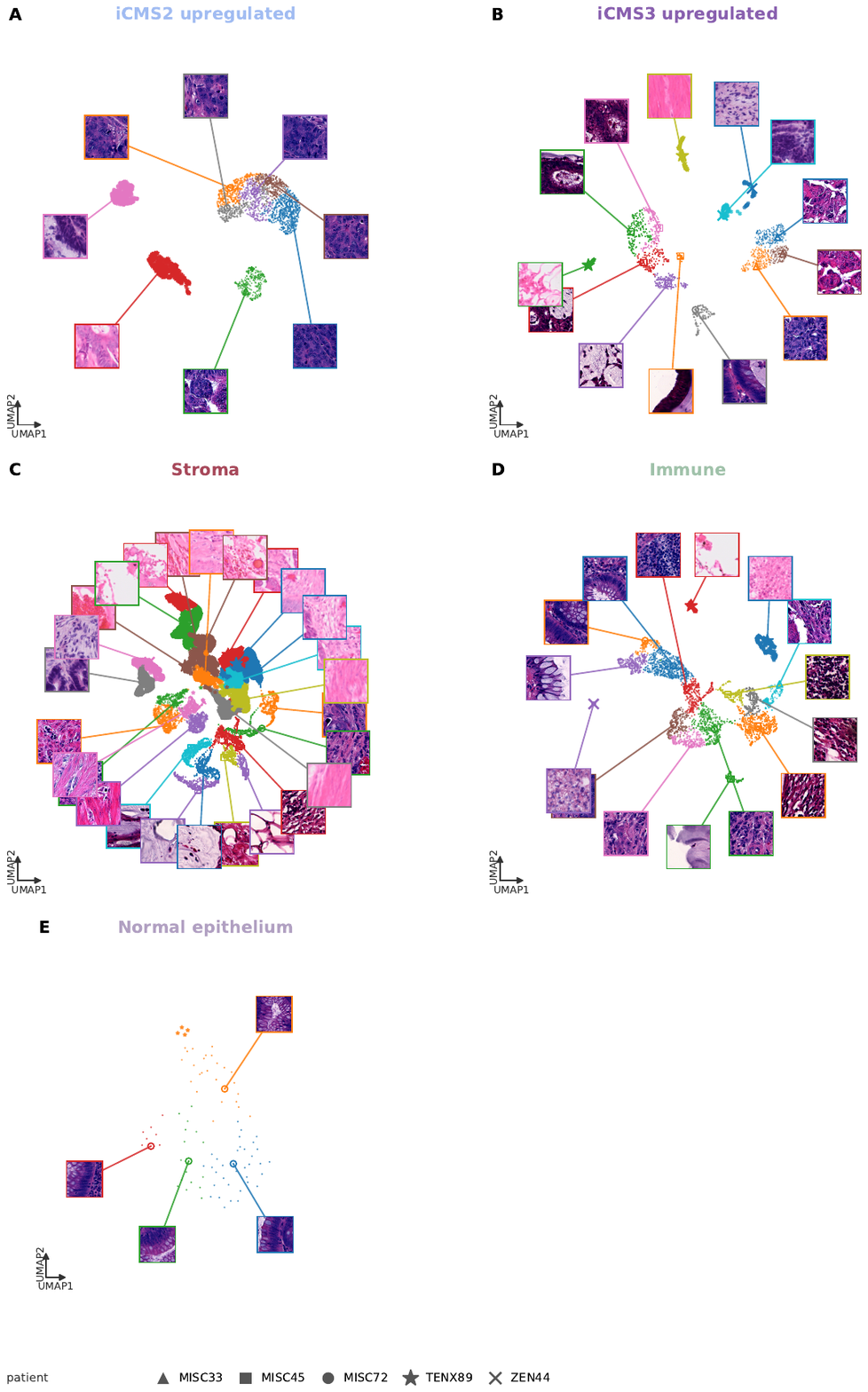}
    \caption{HEST COAD ST data: Relevances of all signatures and colored by their respective prototypes.}
    \label{fig:hest_a4}
\end{figure}

\begin{figure}
    \centering
    \includegraphics[width=0.9\linewidth]{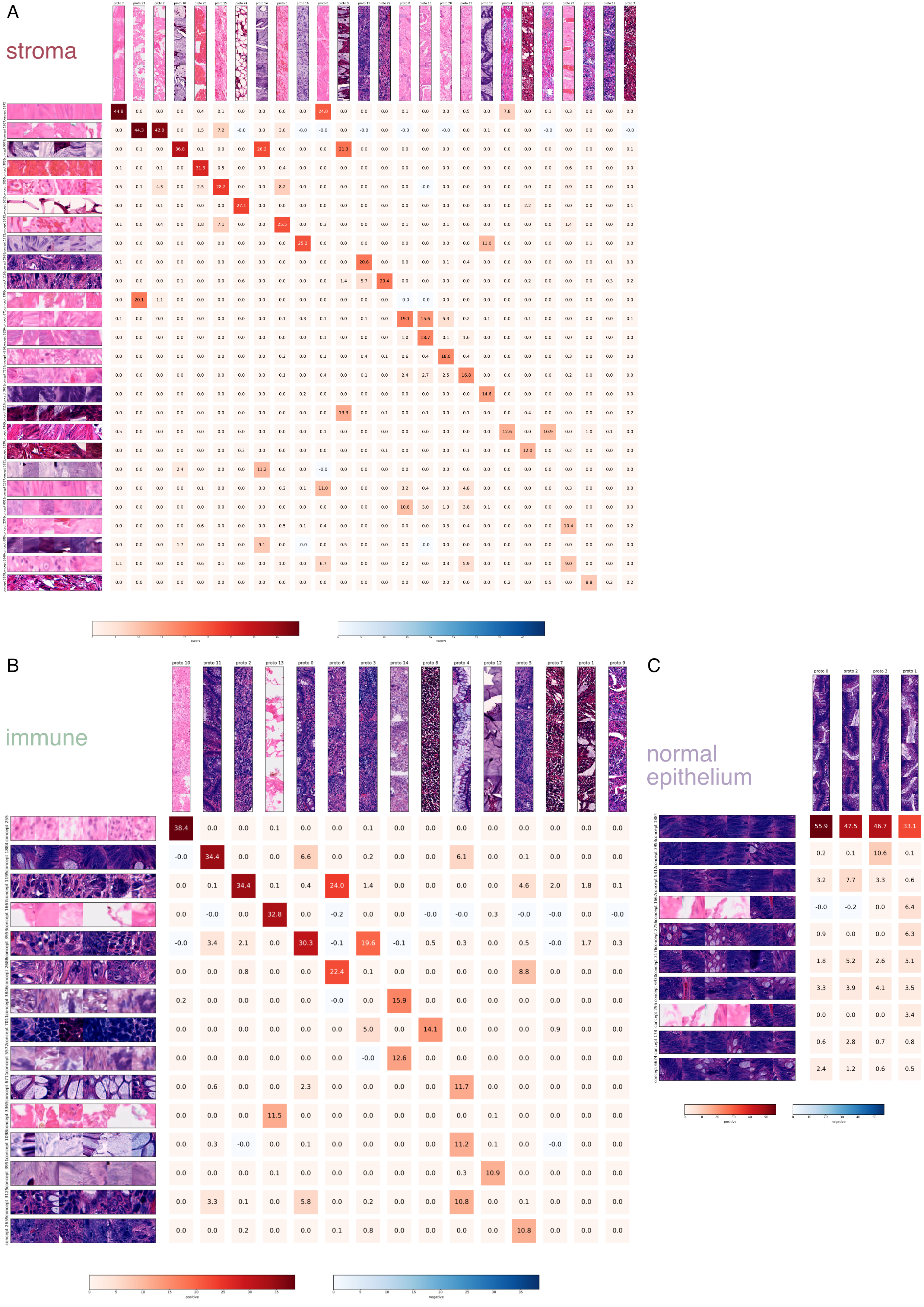}
    \caption{HEST COAD ST data: PCX of stroma, immune and normal epithelium.}
    \label{fig:hest_pcx_st_im_norm}
\end{figure}

\begin{figure}
    \centering
    \includegraphics[width=\linewidth]{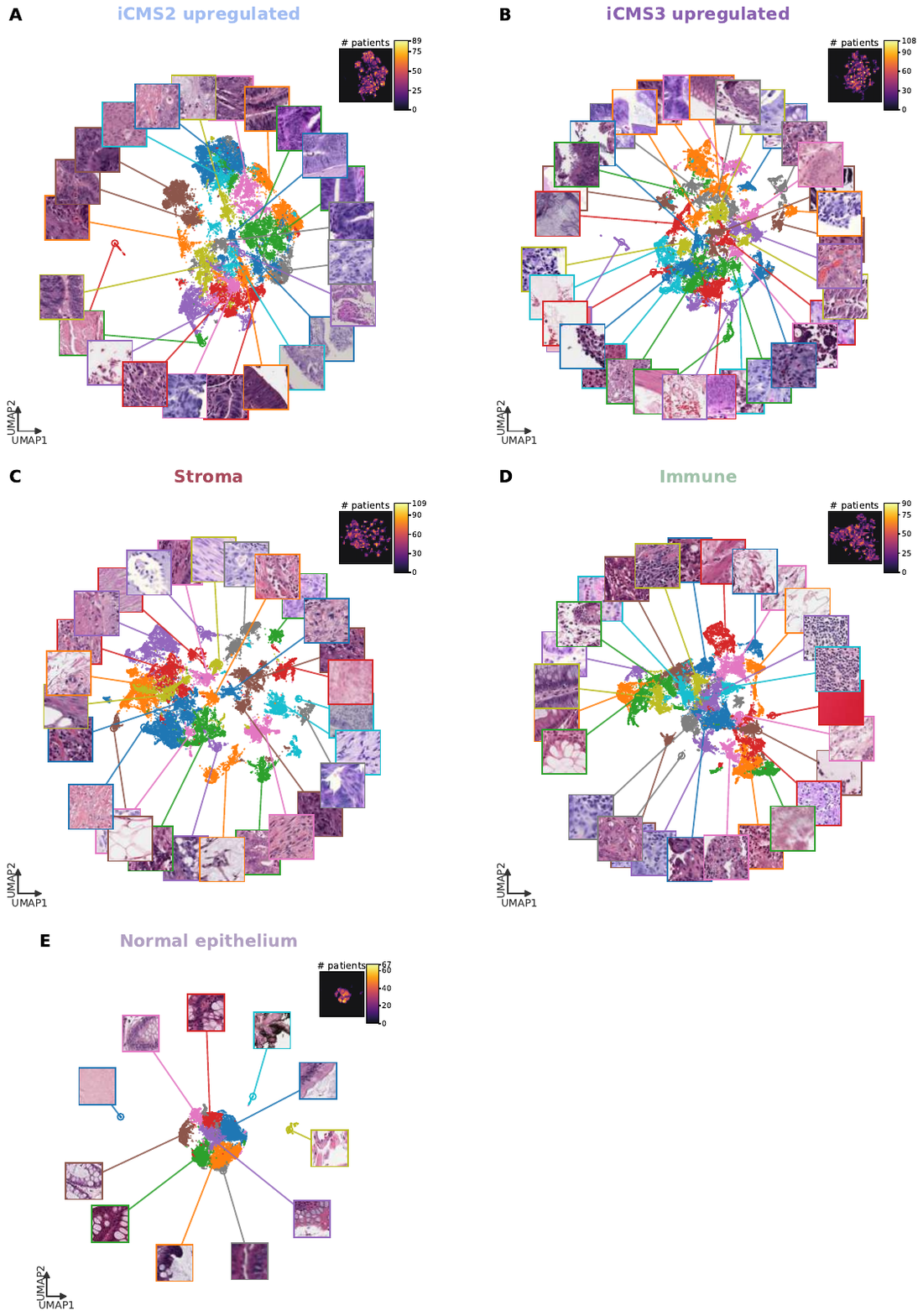}
    \caption{TCGA COAD: Relevances of all signatures and colored by their respective prototypes. Inlets show patient density across prototypes.}
    \label{fig:tcga_a4}
\end{figure}

\begin{figure}
    \centering
    \includegraphics[width=\linewidth]{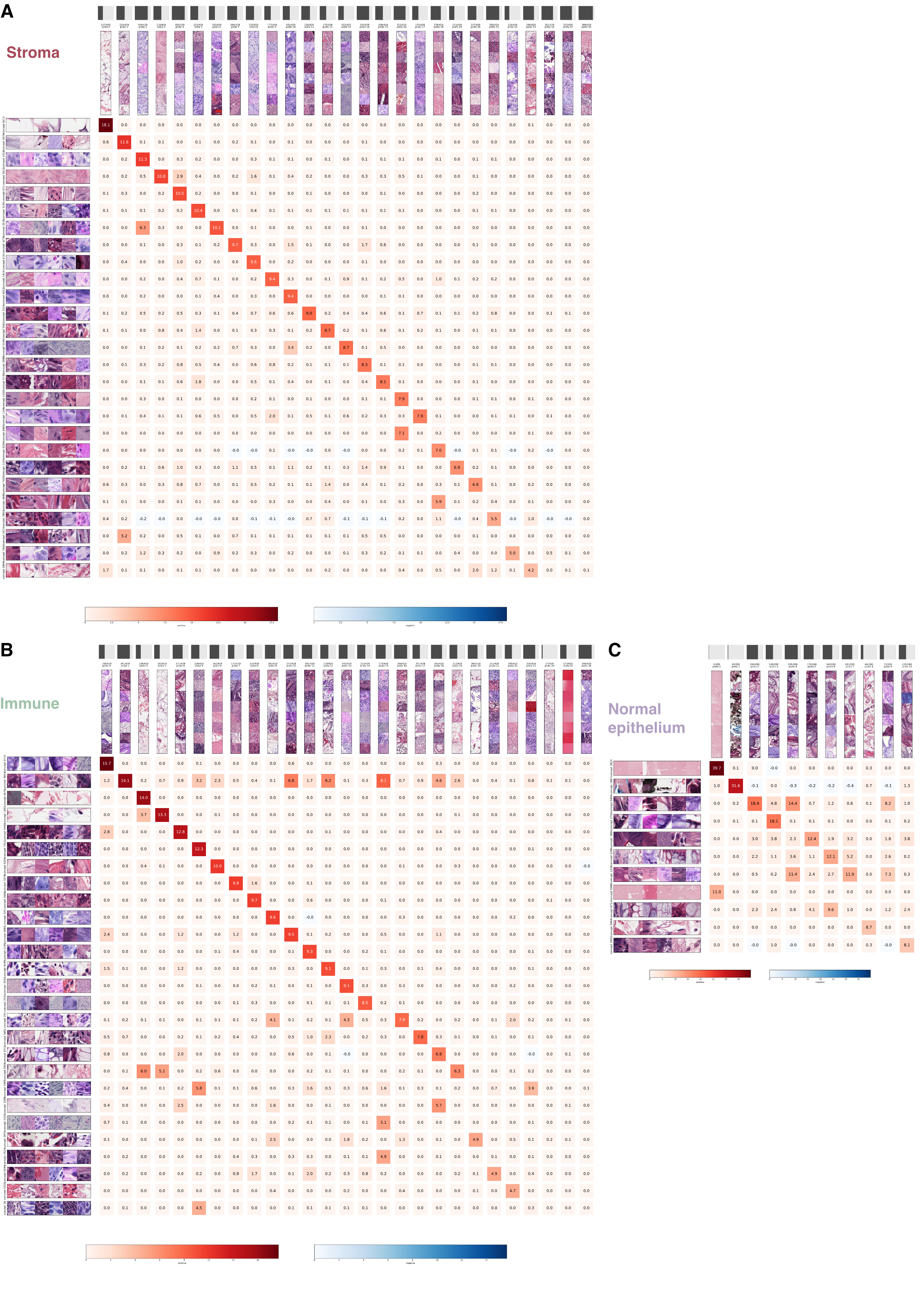}
    \caption{TCGA COAD: Relevance-based concepts of stroma, immune, and normal epithelial with their accompanying prototypes.}
    \label{fig:tcga_pcx_st_im_norm}
\end{figure}

\begin{figure}
\includegraphics[width=1\linewidth]{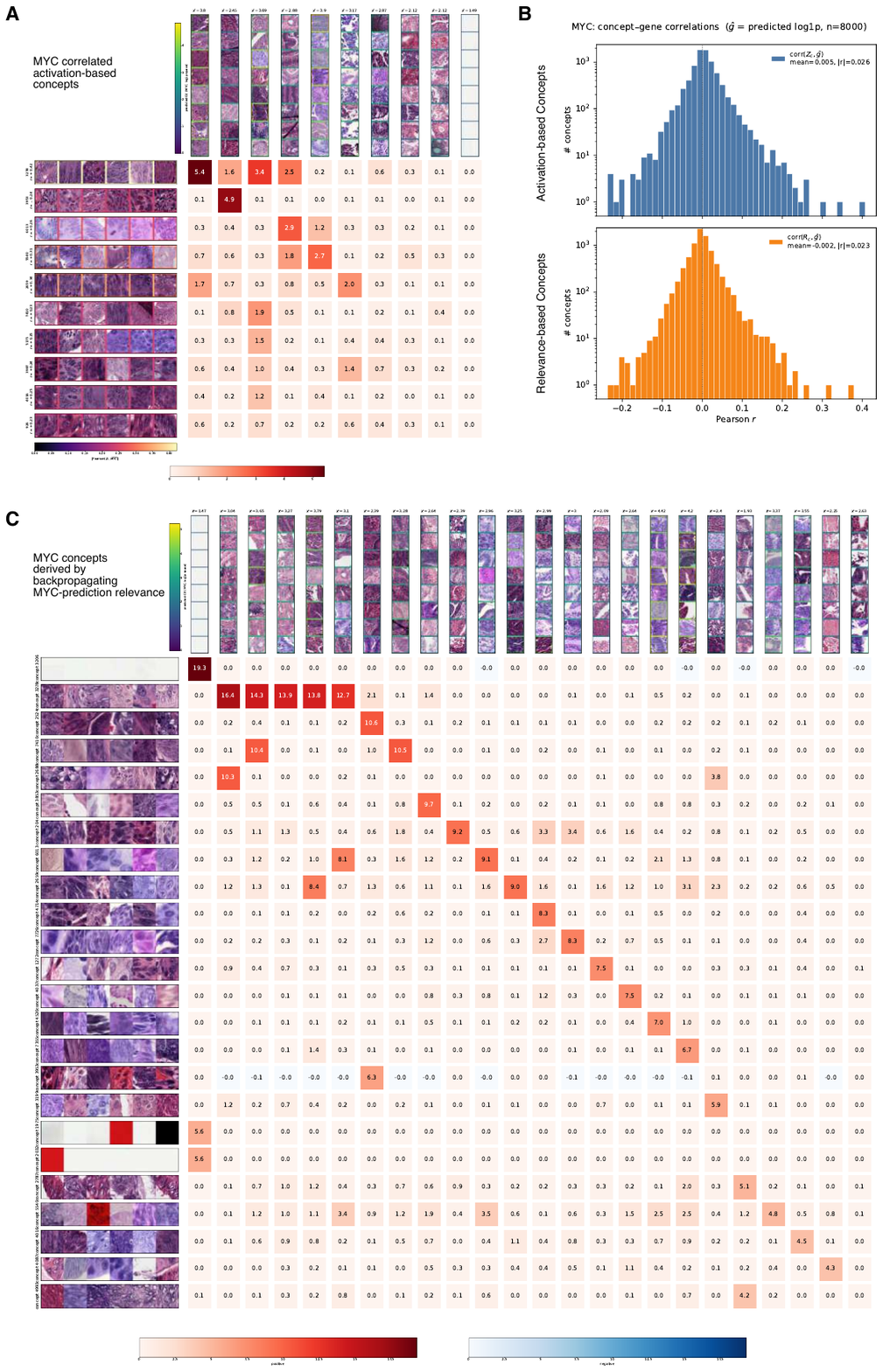} 
\caption{\textbf{Gene-specific concepts based on activations and relevances.} \textbf{A} Activation-based concepts for a single gene (\textit{MYC}). iCMS2 prototypes are displayed, as \textit{MYC} is part of the iCMS2 signature and concepts are selected by maximal correlation with the predicted GE of \textit{MYC}. Concepts were cropped to patch-token activations. \textbf{B} Correlation of all activation- and relevance-based concepts with \textit{MYC} log1p gene expression. \textbf{C} Relevance-based concepts for (\textit{MYC}) explanations. Crops to concepts using LRP to input images.
}
\label{fig:gene_act_rel}
\end{figure}
\clearpage
\bibliographystyle{plainnat}
\bibliography{references}
\end{document}